\pdfoutput=1

\documentclass[11pt]{article}

\usepackage[]{acl}

\usepackage{times}
\usepackage{latexsym}

\usepackage[T1]{fontenc}
\usepackage{CJKutf8}

\usepackage[utf8]{inputenc}

\usepackage{microtype}

\usepackage{inconsolata}

\usepackage{multirow}
\usepackage[T5]{fontenc}

\usepackage{listings}
\usepackage{tabularx}
\usepackage{ltablex}
\usepackage{longtable}
\usepackage{graphicx}
\usepackage{colortbl}
\usepackage{xspace}

\usepackage{pythonhighlight}

\usepackage{supertabular}
\usepackage{xltabular}
\usepackage{amsmath}
\usepackage{amssymb}
\usepackage{mathrsfs}
\usepackage{bm,dutchcal}
\usepackage{float}
\usepackage{tcolorbox}
\usepackage{tipx}
\usepackage{marvosym}

\usepackage{adjustbox}
\usepackage{lineno}
\usepackage{array}
\usepackage{booktabs} 
\usepackage[normalem]{ulem}
\usepackage{booktabs,siunitx}
\usepackage{hhline}
\usepackage{microtype}
\usepackage{longtable}
\usepackage{fancyhdr}

\DeclareRobustCommand{\huggingface}{%
  \begingroup\normalfont
  \vspace{-0.2em}%
  \raisebox{-0.4em}{%
  \includegraphics[height=1.5em]{figures/huggingface_logo.png}%
  }%
  \kern 0.4em%
  \endgroup
}

\DeclareRobustCommand{\github}{%
  \begingroup\normalfont
  \vspace{0.5em}%
  \raisebox{-0.2em}{%
  \includegraphics[height=1.2em]{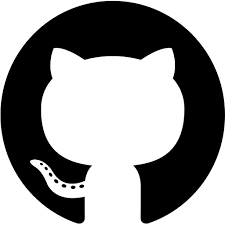}%
  }%
  \kern 0.4em%
  \endgroup
}

\definecolor{custom_light_blue}{rgb}{0.85, 0.95, 1}
\definecolor{custom_light_pink}{rgb}{1, 0.85, 0.85}

\title{VietMed: A Dataset and Benchmark for Automatic  Speech Recognition\\ of Vietnamese in the Medical Domain}

\author{Khai Le-Duc$^{*}$\\
University of Toronto, Canada
\\\texttt{duckhai.le@mail.utoronto.ca}\\
\github \texttt{\href{https://github.com/leduckhai/MultiMed/tree/master/VietMed}{https://github.com/leduckhai/MultiMed/tree/master/VietMed}}\\}

\begin{document}
\maketitle
\begin{abstract}
Due to privacy restrictions, there's a shortage of publicly available speech recognition datasets in the medical domain.
In this work, we present \textit{VietMed} - a Vietnamese speech recognition dataset in the medical domain comprising 16h of labeled medical speech, 1000h of unlabeled medical speech and 1200h of unlabeled general-domain speech.
To our best knowledge, \textit{VietMed} is by far the world's largest public medical speech recognition dataset in 7 aspects: total duration, number of speakers, diseases, recording conditions, speaker roles, unique medical terms and accents.
\textit{VietMed} is also by far the largest public Vietnamese speech dataset in terms of total duration.
Additionally, we are the first to present a medical ASR dataset covering all ICD-10 disease groups and all accents within a country.
Moreover, we release the first public large-scale pre-trained models for Vietnamese ASR, \textit{w2v2-Viet} and \textit{XLSR-53-Viet}, along with the first public large-scale fine-tuned models for medical ASR.
Even without any medical data in unsupervised pre-training, our best pre-trained model \textit{XLSR-53-Viet} generalizes very well to the medical domain by outperforming state-of-the-art \textit{XLSR-53}, from  51.8\% to 29.6\% WER on test set (a relative reduction of more than 40\%).
All code, data and models are made publicly available \href{https://github.com/leduckhai/MultiMed/tree/master/VietMed}{here}.
\end{abstract}

\def\thefootnote{(*)}\footnotetext{Work done during the bachelor thesis at Lehrstuhl Informatik 6 - Machine Learning and Human Language Technology Group, RWTH Aachen University, Germany}\def\thefootnote{\arabic{footnote}}

\thispagestyle{plain}
\pagestyle{plain}

\section{Introduction}

Machine learning models require large amounts of training data.
However, the scarcity of language resources for Vietnamese and especially for the medical domain has been hindering the advancement of corresponding automatic speech recognition (ASR) systems.
Also, the lack of publicly available speech datasets and models in these domains has led to difficulties in reproducing experiments.

Recently, research efforts have been directed towards ASR tasks in the medical field, such as the works \cite{luescher2022:hykist, vieting2023efficient} focused on the development of hybrid ASR systems to transcribe multilingual telephone speech data from patient-physician conversations.
Besides, the works \cite{edwards2017medical, chiu18_interspeech} tackled difficult acoustic conditions and the absence of domain-specific data. 
Nevertheless, none of these studies released their own datasets or pre-trained models.

Out of the limited number of public medical speech datasets we identified, to the best of our knowledge, one of them offers a total of 8 hours of English speech data; however, the dataset's quality is low, as indicated by the authors on their webpage\footnote{https://www.kaggle.com/datasets/paultimothymooney/medical-speech-transcription-and-intent}, where they mentioned issues such as incorrect labels and audio files. 
The second public English medical speech dataset \cite{fareez2022dataset} comprises simulated data, with a predominant focus on respiratory diseases.
This situation restricts investigations to a single disease topic, hindering researchers from exploring experiments related to other medical conditions. 
Also, as pointed out by the authors, this dataset collected speech exclusively from the West England population, which might hurt generalizability to other accents.

Regarding Vietnamese ASR, to the best of our knowledge, there are currently no public large-scale pre-trained models that are peer-reviewed and reproducible\footnote{Several pre-trained models for Vietnamese ASR are available on HuggingFace and GitHub, but none of them have undergone peer review. Their results are self-reported, and we were unable to reproduce them.}.
The \textit{XLSR-53} model \cite{conneau21_interspeech}, was unsupervised pre-trained on 56k hours of 53 languages, but it includes only 200 hours of Vietnamese data. 
Therefore, the constrained performance when fine-tuning the \textit{XLSR-53} model on Vietnamese is conceivable \cite{bachelorthesis}. 

To handle the concerns above, we present a high-quality dataset for Vietnamese medical speech recognition. 
To the best of our knowledge, \textit{VietMed} is by far the world's largest public medical speech dataset in terms of total duration, number of speakers, diseases, recording conditions, speaker roles, unique medical terms and accents. 
Also, \textit{VietMed} is by far the largest public Vietnamese speech dataset in terms of total duration.
Moreover, \textit{VietMed} is the first medical ASR dataset covering all ICD-10 disease groups and all accents within a country.
We then empirically evaluate baseline models on our dataset. 
Our key contributions are:
\begin{itemize}
    \item We present \textit{VietMed} dataset, which includes 16 hours of labeled medical speech, 1000 hours of unlabeled medical speech and 1200 hours of unlabeled general-domain speech.
    \item We release the first public large-scale pre-trained models for Vietnamese ASR, which are peer-reviewed and reproducible.   
    \item We release the first public large-scale fine-tuned models for medical ASR. 
\end{itemize}

Given the transferability of medical terms across languages at some degree, our aim is to contribute to future research in medical ASR for other languages.
All code, data and models are published online\footnote{\url{https://github.com/leduckhai/MultiMed/tree/master/VietMed}}.

\section{Data}
\textit{\textit{VietMed}} data comprises of 3 sets, namely \textit{VietMed-L} for labeled medical speech, \textit{VietMed-U} for unlabeled medical speech, and \textit{Viet-U} for unlabeled general domain speech.
We then split \textit{VietMed-L} into 3 subsets, train (\textit{VietMed-Train}), dev (\textit{VietMed-Dev}) and test (\textit{VietMed-Test}) with duration being 5 hours, 5 hours, and 6 hours respectively, avoiding speaker overlap between the train, dev and test sets.

\subsection{Metadata}
\begin{table}[!ht]
\centering
\resizebox{\columnwidth}{!}{%
\begin{tabular}{|c|c|c|c|} 
\hline
Audio name & Rec.    & Role   & Accent  \\ 
\hline
VietMed\_001       & Tel.    & Doctor & North   \\ 
\hhline{|====|}
Speaker ID & ICD-10  & Gender & Hours   \\ 
\hline
VietMed\_001\_a    & J00-J99 & Male   & 0.06    \\
\hline
\end{tabular}
}
\caption{\label{meta_data} Example of Metadata\_labeled.xlsx. \textit{Rec.} stands for \textit{Recording condition}, in this example is \textit{Tel.} (\textit{Telephone}). Details of ICD-10 codes are shown in Table \ref{icd-10_description} of the Appendix. The speaker role is defined by common roles of speakers in conversations, which typically are: doctor, patient, host, broadcaster, etc.}
\end{table}

We saved all the metadata information to files named Metadata\_labeled.xlsx and Medical\_terms.txt.
As shown in Table \ref{meta_data}, we designed metadata in a way that can support multiple tasks apart from ASR, for example: speaker recognition, keyword recognition, or accent recognition.

\subsection{Data Collection}
We first legally crawled audio data from YouTube under Fair Use Policies\footnote{https://support.google.com/youtube/answer/9783148}$^{,}$\footnote{https://www.copyright.gov/fair-use/} (Details of Fair Use and Consent are in the Appendix).
We manually removed non-speech elements like music, noise, long silences, and any parts that could reveal speaker identities.
Specifically, we removed speaker names, locations where they live, organizations where they work, personal contacts (phone numbers, emails, etc.), personal identifier (date of birth, bank account, id number, etc.), etc.
We converted MP3 audio files to 8kHz wav format, creating 10-30 second segments for \textit{VietMed-U} and \textit{Viet-U}, and <10 second segments for \textit{VietMed-L}. 
Also, we encoded segment names, retaining only ICD-10 code tags to enhance privacy.
Finally, we shuffled all segments of \textit{VietMed-U} and \textit{Viet-U}, making about 500k meaningless segments. 
The purpose here is to prevent immoral users from concatenating segments into meaningful conversations to learn more about speakers.

\subsection{Annotation Process}
Manual annotation of medical spontaneous speech is challenging for humans \cite{edwards2017medical}. 
Annotators may produce varying transcripts. 
Also, applying the fully automated approach \cite{gigaspeech2021} requires large-scale ASR models, which are unavailable in the medical domain and suffer from low quality due to limited human supervision.
We therefore implemented a computer-assisted workflow for medical annotation, outlined as follows:
\begin{enumerate}
    \item We initially gathered transcripts generated by YouTube.
    \item A native Vietnamese with a Biomedical Engineering degree corrected the automatically generated transcripts manually.
    This reduced annotation time by 70\% and improved transcript quality, as it could address issues like stuttering words and speaking rate variations common in real-world conversations.
    \item Another native Vietnamese independently annotated using the same approach.
    \item The resulting two computer-assisted annotation versions were merged and compared. Segments with large differences were excluded.
    \item Finally, we divided the merged transcripts into 3 small validation subsets, where three other Vietnamese with medical backgrounds assessed quality  through manual annotation without assistance by automatic transcription. 
    We then merged the computer-assisted and non-computer-assisted versions as in step 4.
\end{enumerate}

Detailed concerns about the noisy speech in our dataset are shown in the Appendix.

\subsection{Data Statistics}

\begin{table}[!ht]
\centering
\resizebox{\columnwidth}{!}{%
\begin{tabular}{|c|ccc|}
\hline
\multirow{2}{*}{}  & \multicolumn{1}{c|}{Labeled} & \multicolumn{2}{c|}{Unlabeled}      \\ \cline{2-4} 
                   & \multicolumn{2}{c|}{Medical}                             & General \\ \hline
Length {[}hours{]} & \multicolumn{1}{c|}{\texttt{}16}      & \multicolumn{1}{c|}{\texttt{}966}  & 1204    \\ \hline
\#Speakers         & \multicolumn{1}{c|}{\texttt{}61}      & \multicolumn{1}{c|}{2352} & {\texttt{}202}     \\ \hline
\#Record. cond.    & \multicolumn{1}{c|}{\texttt{}\texttt{}8}       & \multicolumn{1}{c|}{\texttt{}\texttt{}\texttt{}9}    & {\texttt{}\texttt{}\texttt{}1}       \\ \hline
\#Med. terms       & \multicolumn{1}{c|}{978}     & \multicolumn{1}{c|}{-}    & -       \\ \hline
\#Accents          & \multicolumn{1}{c|}{\texttt{}\texttt{}6}       & \multicolumn{1}{c|}{\texttt{}\texttt{}\texttt{}6}    & {\texttt{}\texttt{}\texttt{}2}       \\ \hline
\#Roles            & \multicolumn{1}{c|}{\texttt{}\texttt{}6}       & \multicolumn{1}{c|}{\texttt{}\texttt{}\texttt{}6}    & {\texttt{}\texttt{}\texttt{}2}       \\ \hline
\end{tabular}
}
\caption{\label{data_stats} Statistics of \textit{VietMed-L}, \textit{VietMed-U}, \textit{Viet-U}, retrieved from file "Metadata" in the dataset.}
\end{table}

\subsubsection{Labeled Medical Data \textit{VietMed-L}}
In Table \ref{data_stats}, \textit{VietMed-L} contains 16 hours of annotated audio, surpassing other private medical ASR datasets \cite{medisco, chung-etal-2021-data}.
Also, \textit{VietMed-L} has a much higher number of speakers and unique medical terms.
Unlike most datasets that only use simulated scenarios \cite{luescher2022:hykist, fareez2022dataset}, \textit{VietMed-L} captures real-life situations across 8 recording conditions, including telephone (e.g. telemedicine), lectures (e.g. in university hospitals), news (e.g. in medical centers), audiobooks (e.g. medical textbooks), where 85\% of the content is spontaneous speech.
Additionally, we include speech from various roles such as lecturers, hosts, broadcasters, beyond just doctors and patients. 
Furthermore, we ensure diversity by gathering 6 accents representing all regions.

\begin{figure}[!ht]
\begin{center}
\includegraphics[width=\columnwidth]{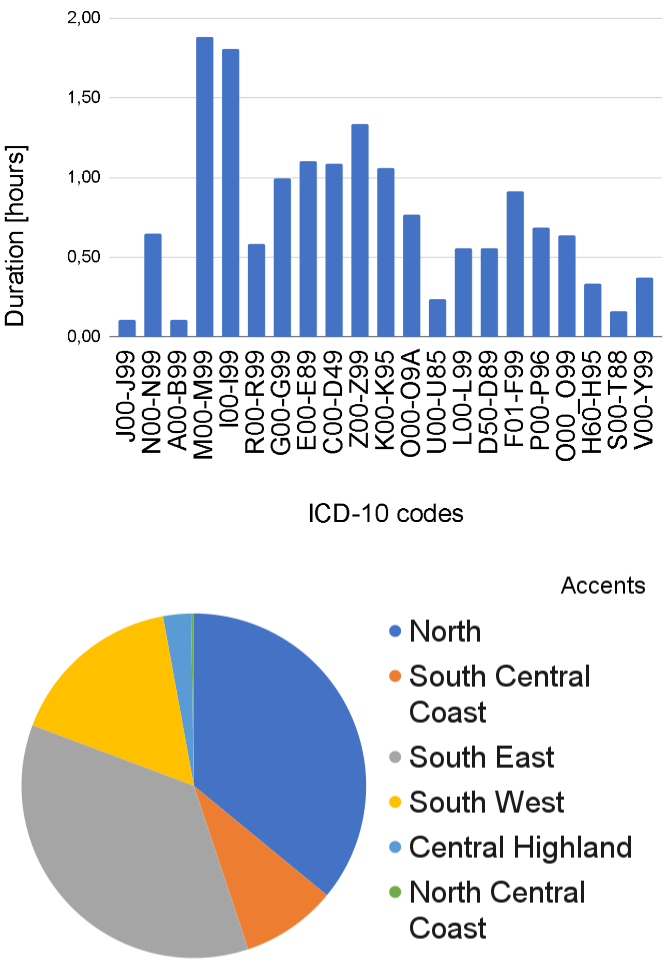} 
\caption{Distribution of ICD-10 codes and accents in \textit{VietMed-L}.}
\label{icd_accents_gender_labeled_medical}
\end{center}
\end{figure}

In Figure \ref{icd_accents_gender_labeled_medical}, rather than primarily focusing on the respiratory disease group (J00-J99) as in \cite{fareez2022dataset}, \textit{VietMed-L} has data from 22/22 disease groups as per World Health Organization (WHO)'s ICD-10 code\footnote{https://www.icd10data.com/ICD10CM/Codes}, supporting the dataset's generalizability.
Also, the accents closely match the real accent  distribution\footnote{https://www.gso.gov.vn/en/population/} (see Table \ref{list_of_regions_vietnam} in the Appendix), and the male/female ratio (54.7\%-45.3\%) is quite balanced.

\subsubsection{Unlabeled Medical Data \textit{VietMed-U}}
In Table \ref{data_stats}, we collected \textit{VietMed-U} in a manner similar to \textit{VietMed-L}, assuring a comparable generalizability as in Figure \ref{icd_accents_gender_labeled_medical}.
Distribution of ICD-10 codes and accents is in Figure \ref{icd_unlabeled_medical} and Figure \ref{accents_unlabeled_medical} of the Appendix.

\subsubsection{Unlabeled General Domain Data \textit{Viet-U}}
\begin{table}[!ht]
\begin{center}
\begin{tabular}{|c|c|} 
\hline
Northern Male   & Southern Male    \\ 
\hline
213h                    & 183h             \\ 
\hhline{|==|}
Northern Female & Southern Female  \\ 
\hline
518h                    & 290h             \\
\hline
\end{tabular}
\caption{\label{unlabeled_general_stats} Genders and accents in \textit{Viet-U}.}
\end{center}
\end{table}

In real world, audiobooks are typically recorded using major Northern and Southern accents.
In Table \ref{unlabeled_general_stats}, statistics of \textit{Viet-U} is shown.

\begin{table*}[!t]
\resizebox{\textwidth}{!}{
\centering
\begin{tabular}{|cc|cc|cc|cc|}
\hline
\multicolumn{2}{|c|}{Trained lexicon}                                                                      & \multicolumn{2}{c|}{LM}                                                                                            & \multicolumn{2}{c|}{\textit{VietMed-Dev}}                                   & \multicolumn{2}{c|}{\textit{VietMed-Test}}                                  \\ \hline
\multicolumn{1}{|c|}{\#words}                                                      & \#vocab               & \multicolumn{1}{c|}{\#words}                                                                       & Size {[}MB{]} & \multicolumn{1}{c|}{OOV}                     & PPL         & \multicolumn{1}{c|}{OOV}                     & PPL         \\ \hline
\multicolumn{1}{|c|}{\multirow{2}{*}{\textit{VietMed-Train} (70k)}}                                 & \multirow{2}{*}{\texttt{}5295} & \multicolumn{1}{c|}{\textit{VietMed-Train} (70k)}                                                                   & {\texttt{}\texttt{}1}             & \multicolumn{1}{c|}{\multirow{2}{*}{0.76\%}} & 149         & \multicolumn{1}{c|}{\multirow{2}{*}{0.66\%}} & 210         \\ \cline{3-4} \cline{6-6} \cline{8-8} 
\multicolumn{1}{|c|}{}                                                             &                       & \multicolumn{1}{c|}{\multirow{2}{*}{\begin{tabular}[c]{@{}c@{}}\textit{VietMed-Train} \\+ \textit{ExtraText}  (8.5M)\end{tabular}}} & {\texttt{}\textbf{98}}   & \multicolumn{1}{c|}{}                        & {\texttt{}\textbf{66}} & \multicolumn{1}{c|}{}                        & {\texttt{}\textbf{84}} \\ \cline{1-2} \cline{4-8} 
\multicolumn{1}{|c|}{\begin{tabular}[c]{@{}c@{}}\textit{VietMed-Train} \\+ \textit{ExtraText}  (8.5M)\end{tabular}} & 33904                 & \multicolumn{1}{c|}{}                                                                              & 103           & \multicolumn{1}{c|}{-}                       & {\texttt{}69}          & \multicolumn{1}{c|}{-}                       & {\texttt{}87}          \\ \hline
\end{tabular}}
\caption{\label{LM_results} Results of 4-gram LMs for 2 lexica.}
\end{table*}

\subsection{Extra Text Data \textit{ExtraText}}
In Table \ref{LM_results}, besides \textit{VietMed-Train} for language model (LM), we used extra text data \textit{ExtraText} to gain lower PPLs. 
Sources are: 
VIVOS\footnote{http://ailab.hcmus.edu.vn/vivos} \cite{vivos_dataset}, BABEL\footnote{https://www.iarpa.gov/research-programs/babel}, CommonVoice\footnote{https://commonvoice.mozilla.org/} \cite{commonvoice_dataset}, FOSD\footnote{https://www.kaggle.com/datasets/thinh127/fpt-open-speech-dataset-fosd-vietnamese} \cite{FOSD_vietnamese_dataset}, VNTC-Health\footnote{https://github.com/duyvuleo/VNTC}, VLSP 2020\footnote{https://vlsp.org.vn/}, ViHealthBERT-FAQ \cite{vihealthbert} and PhoNER-Covid19 \cite{PhoNER_COVID19}.

\subsection{Lexicon}
We used the BABEL project's seed lexicon and augmented it with either \textit{VietMed-Train} or \textit{VietMed-Train} + \textit{ExtraText}.
Using the toolkit Sequitur Grapheme-To-Phoneme\footnote{https://github.com/sequitur-g2p/sequitur-g2p} \cite{G2P_toolkit} - the conversion tool on these pronunciation lexica, the seed lexicon was extended, creating the lexica for training.

\section{Experimental Setups}
For language modelling and initial Gaussian Mixture - Hidden Markov Model (GM-HMM), we followed the same setups and hyperparameters as in \cite{luescher2022:hykist}.
The acoustic model labels were generalized triphone states obtained by classification and regression trees with 4501 labels.
For unsupervised wav2vec 2.0 training \cite{facebook2020wav2vec2} and fine-tuning, we used the same vanilla setups and hyperparameters in \cite{bachelorthesis}.
All models had 118M parameters including 7 CNN layers and 8 Transformer layers.
The last CNN layer had a stride halved for the 8kHz data.
We then chose the pre-training epoch to fine-tune with Framewise Cross-Entropy (fCE) loss that led to the best WERs on dev.
The SpecAugment \cite{park2019specaugment} was used during 33 fine-tuning epochs.

We used RETURNN \cite{zeyer2018returnn} for supervised training and Fairseq \cite{facebook2019fairseq} for unsupervised wav2vec 2.0 training. 
Decoding was performed with RASR \cite{rybach2011rasr}. 
Fairseq models were converted to RETURNN models with our PyTorch-to-RETURNN toolkit\footnote{https://github.com/rwth-i6/pytorch-to-returnn}.

\section{Experimental Results}

\subsection{Language Model}
In Table \ref{LM_results}, augmenting the seed lexicon with only \textit{VietMed-Train} to train \textit{VietMed-Train}+\textit{ExtraText} for LM yields the best PPLs.

\subsection{GM-HMM Alignments}
\begin{table}[!ht]
\begin{center}
\begin{tabular}{|ccccc|}
\hline
\multicolumn{5}{|c|}{WER {[}\%{]} on \textit{VietMed-Dev}}                                                                                 \\ \hline
\multicolumn{1}{|c|}{Mono} & \multicolumn{1}{c|}{Tri}  & \multicolumn{1}{c|}{SAT}  & \multicolumn{1}{c|}{VTLN} & SAT+VTLN \\ \hline
\multicolumn{1}{|c|}{71.7} & \multicolumn{1}{c|}{61.3} & \multicolumn{1}{c|}{52.6} & \multicolumn{1}{c|}{61.3} & 52.2     \\ \hline
\end{tabular}
\caption{\label{gmm_hmm_results} Word-Error-Rates (WERs) [\%] 
of GMM-HMM on \textit{VietMed-Dev}. 
Steps go from Monophone, Triphone to Speaker Adaptive Training + Vocal Tract Length Normalization.}
\end{center}
\end{table}
In Table \ref{gmm_hmm_results}, understanding that WER isn't always a precise metric for alignment quality assessment, we found that WER of SAT was quite similar to SAT+VTLN. 
Therefore, we chose SAT alignments as input for hybrid wav2vec 2.0 training to bypass some steps in GM-HMM process.

\subsection{Hybrid wav2vec 2.0 Baselines}
\begin{table}[!ht]
\centering
\begin{tabular}{|c|cc|}
\hline
\multirow{2}{*}{Pre-trained model} & \multicolumn{2}{c|}{WER {[}\%{]}}                  \\ \cline{2-3} 
                                   & \multicolumn{1}{c|}{dev}           & test          \\ \hline
None                               & \multicolumn{2}{c|}{Non-converged}                 \\ \hline
\textit{XLSR-53}                   & \multicolumn{1}{c|}{45.2}          & 51.8          \\ \hline
\textit{w2v2-Viet}                 & \multicolumn{1}{c|}{45.3}          & 49.5          \\ \hline
\textit{XLSR-53-Viet}              & \multicolumn{1}{c|}{\textbf{26.8}} & \textbf{29.6} \\ \hline
\end{tabular}
\caption{\label{dnn_hmm_results} WERs of wav2vec 2.0 baselines on \textit{VietMed-Dev} and \textit{VietMed-Test}. 
\textit{w2v2-Viet} was pre-trained from scratch on \textit{Viet-U}. 
\textit{XLSR-53-Viet} was pre-trained with \textit{XLSR-53} as initialization on \textit{Viet-U}.
All models have the same architecture and hyperparameters.}
\end{table}
As shown in Table \ref{dnn_hmm_results}, training from scratch did not converge, possibly due to the limited 5-hour fine-tuning data.
\textit{XLSR-53} is a state-of-the-art model pre-trained on 56k hours of 53 languages.
Fine-tuning \textit{XLSR-53} on \textit{VietMed-Train} helped reduce  WER from 52.6\% to 45.2\% on \textit{VietMed-Dev}.
Our \textit{w2v2-Viet} model was competitive to \textit{XLSR-53} despite using 46 times less data for pre-training.
We obtained further improvements by applying our \textit{XLSR-53-Viet} model, which reduced WERs to 26.8\% and 29.6\% on dev and test set respectively, equivalent to relative WERR of 41.8\% compared to the  \textit{XLSR-53} model.
In both our models, we didn't adapt the in-domain data \textit{VietMed-U} during the unsupervised pre-training, although we believed doing so could further enhance WERs and we leave it for future work.

\section{Conclusion}
In this work, we present \textit{VietMed} - a medical speech recognition dataset for Vietnamese.
We introduce a high-quality annotation approach for medical ASR dataset that saves 70\% of time.
Also, we outline our work on creating a LM with acceptable PPL and a compact size.
Finally, our best pre-trained model \textit{XLSR-53-Viet} outperforms the vanilla state-of-the-art \textit{XLSR-53} by reducing WERs from 51.8\% to 29.6\% WER on test set (a relative reduction of more than 40\%) without using any medical data in unsupervised pre-training.

\newpage

\section{Acknowledgements}
This work was partially supported by the project HYKIST funded by the German Federal Ministry of Health on the basis of a decision of the German Federal Parliament (Bundestag) under funding ID ZMVI1-2520DAT04A.

To our best knowledge, this is the very first time in history that the world's largest dataset came from Vietnam.
We thank Minh-Nghia Phan, Peter Vieting, Robin Schmitt, Moritz Gunz, Julian Dierkes for their precious assistance in experimental setups.

We also appreciate Christoph Lüscher, Ralf Schlüter, Hermann Ney for their valuable feedback on this work.

\section*{Version Control: October 2026}
The audio-transcript mismatch issue has been fixed by addressing discrepancies caused by human annotation limitations, particularly missing words at the beginning or end of recordings. 

Therefore, the results reported in this paper should be considered for reference only, as they may not be exactly reproducible using the latest version of the dataset due to transcript updates and improvements.

We thank the community for bringing this issue to our attention, which helped us identify and resolve the audio-transcript mismatch. We appreciate their valuable feedback and contributions to improving the quality of the dataset.
\newpage

\section{Bibliographical References}\label{sec:reference}

\bibliography{lrec-coling2024-example}



\appendix

\section{Ethical Statements}
\subsection{Fair Use}
We strictly followed the criteria of Fair Use by The U.S. Copyright Office\footnote{https://www.copyright.gov/fair-use/}, which also applies to YouTube platform.
Section 107 of the Copyright Act provides the statutory framework for determining whether something is a fair use and identifies certain types of uses—such as criticism, comment, news reporting, teaching, scholarship, and research—as examples of activities that may qualify as fair use. 
Section 107 calls for consideration of the following four factors in evaluating a question of fair use:
\begin{itemize}
    \item (1) \textbf{Purpose and character of the use, including whether the use is of a commercial nature or is for nonprofit educational purposes:}
    Courts look at how the party claiming fair use is using the copyrighted work, and are more likely to find that nonprofit educational and noncommercial uses are fair. 
    Additionally, “transformative” uses are more likely to be considered fair. 
    Transformative uses are those that add something new, with a further purpose or different character, and do not substitute for the original use of the work.
    \item (2) \textbf{Nature of the copyrighted work:} This factor analyzes the degree to which the work that was used relates to copyright’s purpose of encouraging creative expression. 
    Thus, using a more creative or imaginative work (such as a novel, movie, or song) is less likely to support a claim of a fair use than using a factual work (such as a technical article or news item). 
    In addition, use of an unpublished work is less likely to be considered fair.
    \item (3) \textbf{Amount and substantiality of the portion used in relation to the copyrighted work as a whole:} 
    Under this factor, courts look at both the quantity and quality of the copyrighted material that was used.
    That said, some courts have found use of an entire work to be fair under certain circumstances. 
    And in other contexts, using even a small amount of a copyrighted work was determined not to be fair because the selection was an important part—or the “heart”—of the work.
    \item (4) \textbf{Effect of the use upon the potential market for or value of the copyrighted work:} 
    Here, courts review whether, and to what extent, the unlicensed use harms the existing or future market for the copyright owner’s original work.
\end{itemize}

According to the law, we assert our defense under the Fair Use doctrine with the help of Fair Use explanation\footnote{https://copyrightalliance.org/faqs/what-is-fair-use/} by copyrightalliance.org and ELRC Report on legal issues in web crawling \footnote{http://www.elra.info/media/filer\_public/2021/02/12/elrc-legal-analysis-webcrawling\_report-v11.pdf} 
by Pawel Kamocki as follows:
\begin{itemize}
    \item (1) Obviously we crawled the data and published only for non-commercial and research purposes.
    \item (1) We did not directly use videos crawled from YouTube. 
    Instead, we transformed them into audio files with a predefined sampling rate.
    Additionally, we divided lengthy audio files, approximately one hour in duration, into shorter segments lasting between 10 to 30 seconds. 
    These segments were then randomly shuffled, making it impossible for users to piece them together to comprehend the entirety of the originally crawled videos.  
    Therefore, our work is transformative and we do not substitute the original use of the crawled videos.
    \item (2) Our medical conversations are factual (non-fiction) and hence qualified as fair.
    \item (2) Videos on YouTube platform are universally accessible around the world, therefore we satisfy the criteria for the copyrighted work's publication status. 
    \item (3) There is no quantitative test to evaluate whether a given use is fair.
    The randomly shuffled 10-30 second segments we have created do not provide the complete context and meaning of each video, thus making them incapable of representing the "heart" of the copyrighted work.
    \item (4) We don't utilize our publicly available data to compete with the copyright owners' business. 
    Furthermore, our 10-30 second segments have no impact on the viewership count on YouTube. 
    As a result, our efforts do not undermine the potential market being pursued by the copyright owners.
\end{itemize}

Besides our work, several similar works exist that involve the extraction of YouTube videos and their conversion into audio files for research and non-commercial intentions, such as GigaSpeech\footnote{https://github.com/SpeechColab/GigaSpeech} (China \& USA), VoxCeleb\footnote{https://www.robots.ox.ac.uk/~vgg/data/voxceleb/} (UK), VoxLingua107\footnote{https://bark.phon.ioc.ee/voxlingua107/} (UK). 

\subsection{Data Consent}
According to the existing law on the data consent, we are allowed to publish research data. 
We describe in short as follows:
\begin{itemize}
    \item First of all, 137/194 countries signed Data Protection and Privacy Legislation Worldwide\footnote{https://unctad.org/page/data-protection-and-privacy-legislation-worldwide} by the United Nations, including USA, EU, Germany, Vietnam. 
    So Vietnamese law on data protection complies with international law, as Article 6 of the Personal Data Protection Act by the Vietnamese government says: “The protection of personal data is carried out in accordance with international treaties to which the Socialist Republic of Vietnam is a member”.
    \item Researchers have the right to freely publish sensitive medical data for research without the consent of the data subject (speakers in speech data), as Article 20, Section 4 says: “The party processing personal data is not required to register for processing sensitive personal data in the case of research purposes.”
    \item Once more, researchers do not need direct or indirect consent from the data subject to publish research papers, as the Article 16 says: “Data deletion will not apply at the request of the data subject in the following cases: Personal data is processed to serve legal requirements, scientific research, and statistics.”
    \item Again, researchers do not need consent, as Article 9 of the European General Data Protection Regulation (GDPR) permits researchers in Member States to publish personal data for scientific research purposes without consent.
    \item Researchers are strongly encouraged to publish research on sensitive medical data, according to Law on Medical Examination and Treatment, Constitution of the Socialist Republic of Vietnam, Article 22: “Practitioners (…) are responsible for updating relevant medical knowledge (...) including (...) c) Publish scientific research (...).” 
    \item In case of unexpected issues during publishing research, researchers are “Protected by the law and not responsible when a medical incident still occurs after complying with regulations.”, as stated in Article 42.
    \item We crawled generated-by-Vietnam data using Vietnamese IP address and a crawler from a Vietnamese company authorized by Vietnamese government, and the right to publish this data for research purposes is protected under Vietnamese Law (shown above), since Google (Youtube) must comply with Vietnamese law on content in Vietnamese cyberspace, as shown in Article 26, Cybersecurity Law, Constitution of the Socialist Republic of Vietnam: “Domestic and foreign enterprises providing services on telecommunications networks, the Internet, and value-added services in cyberspace in Vietnam have activities of collecting, exploiting, analyzing, and processing information data (...) created by service users in Vietnam must store this data in Vietnam (...) as prescribed by the Government.”
    \item International researchers have the right to publish and process Vietnamese personal data without consent. 
    Also they are both encouraged to publish Vietnamese research data and are protected under Vietnamese law because they must comply with Vietnamese law on generated-by-Vietnam data, according to Article 2 and 10, the Vietnamese Civil Code on Civil Relations with Foreign Elements: “The provisions of Vietnamese civil law apply to civil relations involving foreign elements (...). 
    In case the application or consequences of the application of foreign law are contrary to (...) the Vietnam Civil Code and other basic principles of Vietnamese law , then Vietnamese law applies.”
\end{itemize}

The YouTube content in our dataset is about medical shows, interviews, lectures, etc., where all participants talked to camera and were aware that the videos are publicly accessible in an attempt to provide medical knowledge to YouTube users. 
These videos are published by national TV channels, not by some amateur content creators. 
There are some YouTube videos that speakers are not aware of being recorded, published by amateurs, but we did not include them in our dataset.

\section{Additional Details of \textit{VietMed} Dataset}

\subsection{Description of ICD-10 Codes}
\begin{table*}[!t]
\begin{center}
\resizebox{\textwidth}{!}{%
\begin{tabular}{|l|l|}
\hline
\multicolumn{1}{|c|}{\textbf{ICD-10 Code}} & \multicolumn{1}{c|}{\textbf{Description of diseases}}                                               \\ \hline
A00-B99                                    & Certain infectious and parasitic diseases                                                           \\ \hline
C00-D49                                    & Neoplasms                                                                                           \\ \hline
D50-D89                                    & Diseases of the blood and blood-forming organs and certain disorders involving the immune mechanism \\ \hline
E00-E89                                    & Endocrine, nutritional and metabolic diseases                                                       \\ \hline
F01-F99                                    & Mental, Behavioral and Neurodevelopmental disorders                                                 \\ \hline
G00-G99                                    & Diseases of the nervous system                                                                      \\ \hline
H00-H59                                    & Diseases of the eye and adnexa                                                                      \\ \hline
H60-H95                                    & Diseases of the ear and mastoid process                                                             \\ \hline
I00-I99                                    & Diseases of the circulatory system                                                                  \\ \hline
J00-J99                                    & Diseases of the respiratory system                                                                  \\ \hline
K00-K95                                    & Diseases of the digestive system                                                                    \\ \hline
L00-L99                                    & Diseases of the skin and subcutaneous tissue                                                        \\ \hline
M00-M99                                    & Diseases of the musculoskeletal system and connective tissue                                        \\ \hline
N00-N99                                    & Diseases of the genitourinary system                                                                \\ \hline
O00-O9A                                    & Pregnancy, childbirth and the puerperium                                                            \\ \hline
P00-P96                                    & Certain conditions originating in the perinatal period                                              \\ \hline
Q00-Q99                                    & Congenital malformations, deformations and chromosomal abnormalities                                \\ \hline
R00-R99                                    & Symptoms, signs and abnormal clinical and laboratory findings, not elsewhere classified             \\ \hline
S00-T88                                    & Injury, poisoning and certain other consequences of external causes                                 \\ \hline
U00-U85                                    & Codes for special purposes                                                                          \\ \hline
V00-Y99                                    & External causes of morbidity                                                                        \\ \hline
Z00-Z99                                    & Factors influencing health status and contact with health services                                  \\ \hline
\end{tabular}%
}
\caption{\label{icd-10_description} Description of ICD-10 codes which our dataset follows, according to the 2024 version by World Health Organziation. Each ICD-10 Code, e.g. A00-B99, could be in smaller codes partitioned. However, in our dataset we only used 22 ICD-10 Codes since partitioning into smaller codes makes the annotation too complicated and unnecessary.}
\end{center}
\end{table*}
Table \ref{icd-10_description} shows the detailed description of ICD-10 codes.
The audio files in our dataset are classified based on these ICD-10 codes.

\subsection{Real Distribution of Accents in Vietnam}
\begin{table*}[!t]
\begin{center}
\begin{tabular}{|l|l|l|c|}
\hline
\multicolumn{1}{|c|}{\textbf{Region}} & \multicolumn{1}{c|}{\textbf{Subregion}} & \multicolumn{1}{c|}{\textbf{Typical Provinces}}                    & \textbf{Population} \\ \hline
\multirow{3}{*}{North}                & Northest                                & \begin{tabular}[c]{@{}l@{}}Cao Bằng\\ Hà Giang\\ ...\end{tabular}  & 8M                  \\ \cline{2-4} 
                                      & Northwest                               & \begin{tabular}[c]{@{}l@{}}Điện Biên\\ Hòa Bình\\ ...\end{tabular} & 4M                  \\ \cline{2-4} 
                                      & Red River Delta                         & \begin{tabular}[c]{@{}l@{}}Hà Nội \\ Hải Phòng\\ ...\end{tabular}  & 20M                 \\ \hline
\multirow{3}{*}{Central}              & North Central Coast                     & \begin{tabular}[c]{@{}l@{}}Hà Tĩnh\\ Nghệ An\\ ...\end{tabular}    & 10M                 \\ \cline{2-4} 
                                      & South Central Coast                     & \begin{tabular}[c]{@{}l@{}}Đà Nẵng\\ Bình Thuận\\ ...\end{tabular} & 9M                  \\ \cline{2-4} 
                                      & Central Highland                        & \begin{tabular}[c]{@{}l@{}}Gia Lai\\ Kon Tum\\ ...\end{tabular}          & 5M                  \\ \hline
\multirow{2}{*}{South}                & Southeast                               & \begin{tabular}[c]{@{}l@{}}TP. Hồ Chí Minh\\ Đồng Nai\\ ...\end{tabular} & 16M                 \\ \cline{2-4} 
                                      & Southwest                               & \begin{tabular}[c]{@{}l@{}}Long An\\ Cần Thơ\\ ...\end{tabular}    & 18M                 \\ \hline
\end{tabular}
\label{list_of_regions_vietnam}
\caption{Real distribution of Vietnamese accents. The statistics was retrieved in 2015 from Vietnamese General Statistics Office. In our dataset, we did not split the North accent into subregional accents since it was too difficult for our annotators to correctly recognize subregional accents of the North region.}
\end{center}
\end{table*}
Table \ref{list_of_regions_vietnam} shows the real distribution of accents in Vietnam, which our \textit{VietMed} dataset follows.

\subsection{Concerns about Noisy Speech in \textit{VietMed}}
Real-world speech data should contain real-world acoustic conditions (e.g. background noises, music, etc.). 
To enhance the quality of a speech dataset, especially for a read speech dataset, people often use a Signal-to-Noise Ratio (SNR) to measure the background noises and discard segments with a high level of SNR.
However, using an SNR threshold to obtain only good speech signals, discarding noisy segments, would violate real-world scenarios, making our \textit{VietMed} dataset no longer real world but rather "simulated".

Actually, we only removed audio segments that have no speech. 
We still kept overlapped speech segments, as long as the main speaker's speech is still comprehensible. 
The quality assurance for real-world ASR datasets should focus on transcription, which we have already addressed in the paper, instead of focusing on the quality of the speech signal.

\subsection{Extra Data Statistics for Labeled Medical Data \textit{VietMed-L}}
\begin{table*}[!ht]
\begin{center}
\begin{tabular}{|c|c|c|c|}
\hline
                     & \textit{VietMed-Train} & \textit{VietMed-Dev}   & \textit{VietMed-Test}  \\ \hline
Dur. {[}hours{]} & {\texttt{}\texttt{}5}     & {\texttt{}\texttt{}5}     & {\texttt{}\texttt{}6}     \\ \hline
\#Speakers           & {\texttt{}13}    & {\texttt{}21}    & {\texttt{}27}    \\ \hline
\#Words              & 70k & 69k & 76k \\ \hline
\#Rec. cond. & {\texttt{}\texttt{}2}     & {\texttt{}\texttt{}4}     & {\texttt{}\texttt{}6}     \\ \hline
\#Accents            & {\texttt{}\texttt{}3}     & {\texttt{}\texttt{}4}     & {\texttt{}\texttt{}5}     \\ \hline
\#Roles              & {\texttt{}\texttt{}3}     & {\texttt{}\texttt{}4}     & {\texttt{}\texttt{}6}     \\ \hline
\end{tabular}
\caption{\label{data_stats_traindevtest} Data statistics of \textit{VietMed-L}, retrieved from file "Metadata" in the dataset.}
\end{center}
\end{table*}
Table \ref{data_stats_traindevtest} shows the statistics of 3 train-dev-test subsets in \textit{VietMed-L}.
We split these 3 subsets in a way that made \textit{VietMed-Train} the least generalizability by having the least number of speakers, recording conditions, accents and roles, while prioritizing \textit{VietMed-Dev} and \textit{VietMed-Test} more generalizability.
Note that no speaker overlap occured in the 3 subsets.

\subsection{Extra Data Statistics for Unlabeled Medical Data \textit{VietMed-U}}

\begin{figure*}[!ht]
\begin{center}
\includegraphics[scale=0.8]{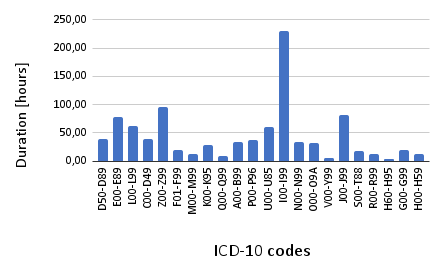} 
\caption{Distribution of ICD-10 code in \textit{VietMed-U}.}
\label{icd_unlabeled_medical}
\end{center}
\end{figure*}

\begin{figure*}[!ht]
\begin{center}
\includegraphics[scale=0.6]{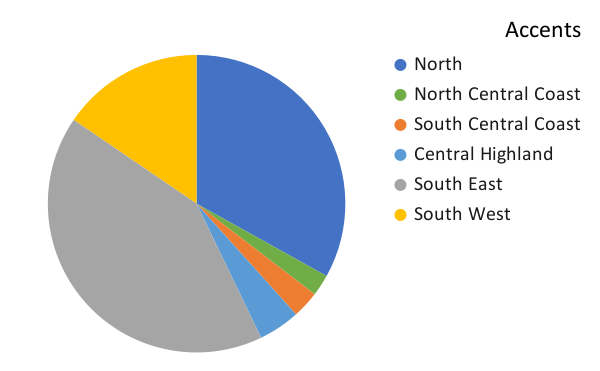} 
\caption{Distribution of accents in \textit{VietMed-U}.}
\label{accents_unlabeled_medical}
\end{center}
\end{figure*}

Figure \ref{icd_unlabeled_medical} shows the distribution of ICD-10 code and Figure \ref{accents_unlabeled_medical} shows the distribution of accents in \textit{VietMed-U}.
We collected \textit{VietMed-U} in a manner similar to \textit{VietMed-L}, assuring a comparable generalizability as in \textit{VietMed-L}.

\section{ASR Error Analysis}

\subsection{Error Analysis of Pre-trained Model}

Table \ref{ASR_erroranalysis_XLSR53_test} shows the error analysis of our pre-trained model \textit{XLSR-53} on the \textit{VietMed-Test} set.

Table \ref{ASR_erroranalysis_w2v2Viet_test} shows the error analysis of our pre-trained model \textit{w2v2-Viet} on the \textit{VietMed-Test} set.

Table \ref{ASR_erroranalysis_XLSR53Viet_test} shows the error analysis of our best pre-trained model \textit{XLSR-53-Viet} on the \textit{VietMed-Test} set.

\subsection{Error Analysis of Confusion Pairs}
Table \ref{confusion_pair} shows the statistics of confusion pairs in \textit{VietMed-Test} using the best pre-trained model \textit{XLSR-53-Viet}.
Closely similar words could lead to the decreased accuracy of an ASR system.
Therefore, collecting confusion pairs which the ASR system often misrecognized gives researchers an opportunity to analyze common ASR errors and improve the ASR accuracy.

As shown in the table, words that are parts of medical terms and fillers contribute greatly to the decreased accuracy of the ASR system using the pre-trained model \textit{XLSR-53-Viet}.
This difficulty was confirmed by our annotators during the dataset annotation, since it was very hard to correctly transcribe medical terms and fillers in real-world medical conversations.

\subsection{Error Analysis of OOV}
Table \ref{OOV_IARPA} shows the list of OOVs loan words found in \textit{VietMed-Train}. 
In this table, we used the BABEL project's seed lexicon and automatically augmented it with \textit{VietMed-Train}.
We used the toolkit Sequitur Grapheme-To-Phoneme\footnote{https://github.com/sequitur-g2p/sequitur-g2p} \cite{G2P_toolkit} - the conversion tool on these pronunciation lexica, to extend the seed lexicon, creating the lexicon for training.

First, we found that the seed lexicon by BABEL was overwhelmed by North and North Central Coast accents, leaving almost no other accents like South Central Coast, Central Highland, Southwest and Southeast.
Therefore, this lexicon hurts the accuracy of ASR systems on a generalized dataset like \textit{VietMed}.
Second, \textit{VietMed} has a very large number of medical terms, which often come from English loan words.
So automatic extension of the seed lexicon without human correction led to wrong phoneme mapping of medical terms, which also hurts the accuracy of ASR systems.

\onecolumn
\newpage

\begin{table*}[h]
\begin{center}
\resizebox{\textwidth}{!}{%
\begin{tabular}{|lccccc|c|c|l|l|l|l|l|l|}
\hline
\multicolumn{1}{|c|}{\textbf{Speaker ID}} & \multicolumn{1}{c|}{\textbf{Rec.}}           & \multicolumn{1}{c|}{\textbf{ICD-10}}          & \multicolumn{1}{c|}{\textbf{Role}} & \multicolumn{1}{c|}{\textbf{Gend}} & \textbf{Acc.} & \textbf{\# Snt} & \textbf{\# Wrd} & \multicolumn{1}{c|}{\textbf{Corr}} & \multicolumn{1}{c|}{\textbf{Sub}} & \multicolumn{1}{c|}{\textbf{Del}} & \multicolumn{1}{c|}{\textbf{Ins}} & \multicolumn{1}{c|}{\textbf{Err}} & \multicolumn{1}{c|}{\textbf{S.Err}} \\ \hline
\multicolumn{1}{|l|}{vietmed\_002}        & \multicolumn{1}{c|}{\multirow{7}{*}{Tel.}}   & \multicolumn{1}{c|}{N00-N99}                  & \multicolumn{1}{c|}{Lec.}          & \multicolumn{1}{c|}{F}             & SCC           & 363             & 7631            & 30.7                               & 54.4                              & 14.9                              & 5.6                               & 74.9                              & 100.0                               \\ \cline{1-1} \cline{3-14} 
\multicolumn{1}{|l|}{vietmed\_004}        & \multicolumn{1}{c|}{}                        & \multicolumn{1}{c|}{M00-M99}                  & \multicolumn{1}{c|}{Doc.}          & \multicolumn{1}{c|}{M}             & SCC           & 446             & 10575           & 51.7                               & 34.8                              & 13.5                              & 6.8                               & 55.0                              & 100.0                               \\ \cline{1-1} \cline{3-14} 
\multicolumn{1}{|l|}{vietmed\_014\_a}     & \multicolumn{1}{c|}{}                        & \multicolumn{1}{c|}{\multirow{2}{*}{K00-K95}} & \multicolumn{1}{c|}{Host}          & \multicolumn{1}{c|}{F}             & N             & 18              & 491             & 63.7                               & 23.4                              & 12.8                              & 3.7                               & 39.9                              & 100.0                               \\ \cline{1-1} \cline{4-14} 
\multicolumn{1}{|l|}{vietmed\_014\_b}     & \multicolumn{1}{c|}{}                        & \multicolumn{1}{c|}{}                         & \multicolumn{1}{c|}{Doc.}          & \multicolumn{1}{c|}{M}             & N             & 164             & 4034            & 59.6                               & 28.5                              & 11.9                              & 5.2                               & 45.6                              & 100.0                               \\ \cline{1-1} \cline{3-14} 
\multicolumn{1}{|l|}{vietmed\_015\_a}     & \multicolumn{1}{c|}{}                        & \multicolumn{1}{c|}{\multirow{3}{*}{O00-O9A}} & \multicolumn{1}{c|}{Host}          & \multicolumn{1}{c|}{F}             & N             & 73              & 1779            & 68.8                               & 20.3                              & 10.9                              & 4.2                               & 35.4                              & 100.0                               \\ \cline{1-1} \cline{4-14} 
\multicolumn{1}{|l|}{vietmed\_015\_b}     & \multicolumn{1}{c|}{}                        & \multicolumn{1}{c|}{}                         & \multicolumn{1}{c|}{Doc.}          & \multicolumn{1}{c|}{F}             & N             & 297             & 5669            & 58.8                               & 28.3                              & 12.9                              & 4.4                               & 45.6                              & 100.0                               \\ \cline{1-1} \cline{4-14} 
\multicolumn{1}{|l|}{vietmed\_015\_c}     & \multicolumn{1}{c|}{}                        & \multicolumn{1}{c|}{}                         & \multicolumn{1}{c|}{Pat.}          & \multicolumn{1}{c|}{F}             & N             & 55              & 1010            & 43.0                               & 37.3                              & 19.7                              & 3.7                               & 60.7                              & 100.0                               \\ \hline
\multicolumn{1}{|l|}{vietmed\_017\_a}     & \multicolumn{1}{c|}{\multirow{10}{*}{Talk.}} & \multicolumn{1}{c|}{\multirow{2}{*}{U00-U85}} & \multicolumn{1}{c|}{Doc.}          & \multicolumn{1}{c|}{F}             & SW            & 47              & 1104            & 50.0                               & 37.2                              & 12.8                              & 5.4                               & 55.4                              & 100.0                               \\ \cline{1-1} \cline{4-14} 
\multicolumn{1}{|l|}{vietmed\_017\_b}     & \multicolumn{1}{c|}{}                        & \multicolumn{1}{c|}{}                         & \multicolumn{1}{c|}{Doc.}          & \multicolumn{1}{c|}{M}             & N             & 86              & 2061            & 62.8                               & 26.9                              & 10.2                              & 5.0                               & 42.2                              & 100.0                               \\ \cline{1-1} \cline{3-14} 
\multicolumn{1}{|l|}{vietmed\_018\_a}     & \multicolumn{1}{c|}{}                        & \multicolumn{1}{c|}{\multirow{6}{*}{K00-K95}} & \multicolumn{1}{c|}{Host}          & \multicolumn{1}{c|}{F}             & SW            & 63              & 1527            & 54.3                               & 32.5                              & 13.2                              & 19.6                              & 65.3                              & 100.0                               \\ \cline{1-1} \cline{4-14} 
\multicolumn{1}{|l|}{vietmed\_018\_b}     & \multicolumn{1}{c|}{}                        & \multicolumn{1}{c|}{}                         & \multicolumn{1}{c|}{Doc.}          & \multicolumn{1}{c|}{M}             & SW            & 192             & 5293            & 59.8                               & 26.2                              & 14.0                              & 7.2                               & 47.4                              & 100.0                               \\ \cline{1-1} \cline{4-14} 
\multicolumn{1}{|l|}{vietmed\_018\_c}     & \multicolumn{1}{c|}{}                        & \multicolumn{1}{c|}{}                         & \multicolumn{1}{c|}{Doc.}          & \multicolumn{1}{c|}{F}             & SW            & 118             & 2761            & 55.3                               & 31.4                              & 13.2                              & 8.7                               & 53.3                              & 100.0                               \\ \cline{1-1} \cline{4-14} 
\multicolumn{1}{|l|}{vietmed\_018\_d}     & \multicolumn{1}{c|}{}                        & \multicolumn{1}{c|}{}                         & \multicolumn{1}{c|}{Pat.}          & \multicolumn{1}{c|}{F}             & SW            & 20              & 412             & 33.3                               & 36.9                              & 29.9                              & 6.1                               & 72.8                              & 100.0                               \\ \cline{1-1} \cline{4-14} 
\multicolumn{1}{|l|}{vietmed\_018\_e}     & \multicolumn{1}{c|}{}                        & \multicolumn{1}{c|}{}                         & \multicolumn{1}{c|}{Pat.}          & \multicolumn{1}{c|}{M}             & SW            & 5               & 76              & 31.6                               & 40.8                              & 27.6                              & 10.5                              & 78.9                              & 100.0                               \\ \cline{1-1} \cline{4-14} 
\multicolumn{1}{|l|}{vietmed\_018\_f}     & \multicolumn{1}{c|}{}                        & \multicolumn{1}{c|}{}                         & \multicolumn{1}{c|}{Doc.}          & \multicolumn{1}{c|}{M}             & SW            & 25              & 639             & 41.2                               & 42.9                              & 16.0                              & 5.0                               & 63.8                              & 100.0                               \\ \cline{1-1} \cline{3-14} 
\multicolumn{1}{|l|}{vietmed\_019\_a}     & \multicolumn{1}{c|}{}                        & \multicolumn{1}{c|}{\multirow{2}{*}{L00-L99}} & \multicolumn{1}{c|}{Host}          & \multicolumn{1}{c|}{F}             & SW            & 58              & 1490            & 55.1                               & 31.9                              & 13.0                              & 6.9                               & 51.8                              & 100.0                               \\ \cline{1-1} \cline{4-14} 
\multicolumn{1}{|l|}{vietmed\_019\_b}     & \multicolumn{1}{c|}{}                        & \multicolumn{1}{c|}{}                         & \multicolumn{1}{c|}{Doc.}          & \multicolumn{1}{c|}{F}             & SW            & 116             & 2776            & 56.5                               & 30.5                              & 13.0                              & 7.7                               & 51.3                              & 100.0                               \\ \hline
\multicolumn{1}{|l|}{vietmed\_023}        & \multicolumn{1}{c|}{\multirow{2}{*}{Pod.}}   & \multicolumn{1}{c|}{P00-P96}                  & \multicolumn{1}{c|}{Pod.}          & \multicolumn{1}{c|}{F}             & SW            & 390             & 7414            & 55.4                               & 35.8                              & 8.8                               & 4.9                               & 49.6                              & 99.7                                \\ \cline{1-1} \cline{3-14} 
\multicolumn{1}{|l|}{vietmed\_024}        & \multicolumn{1}{c|}{}                        & \multicolumn{1}{c|}{O00\_O99}                 & \multicolumn{1}{c|}{Pod.}          & \multicolumn{1}{c|}{F}             & SE            & 376             & 7425            & 61.2                               & 28.8                              & 10.0                              & 4.7                               & 43.5                              & 99.7                                \\ \hline
\multicolumn{1}{|l|}{vietmed\_025\_a}     & \multicolumn{1}{c|}{\multirow{2}{*}{Diag.}}  & \multicolumn{1}{c|}{\multirow{2}{*}{H60-H95}} & \multicolumn{1}{c|}{Host}          & \multicolumn{1}{c|}{F}             & SW            & 101             & 2280            & 60.3                               & 29.1                              & 10.7                              & 5.0                               & 44.7                              & 100.0                               \\ \cline{1-1} \cline{4-14} 
\multicolumn{1}{|l|}{vietmed\_025\_b}     & \multicolumn{1}{c|}{}                        & \multicolumn{1}{c|}{}                         & \multicolumn{1}{c|}{Doc.}          & \multicolumn{1}{c|}{M}             & SE            & 91              & 1838            & 65.7                               & 24.8                              & 9.5                               & 6.6                               & 40.9                              & 100.0                               \\ \hline
\multicolumn{1}{|l|}{vietmed\_026}        & \multicolumn{1}{c|}{Lec.}                    & \multicolumn{1}{c|}{A00-B99}                  & \multicolumn{1}{c|}{Lec.}          & \multicolumn{1}{c|}{M}             & NCC           & 21              & 355             & 31.8                               & 47.6                              & 20.6                              & 6.5                               & 74.6                              & 100.0                               \\ \hline
\multicolumn{1}{|l|}{vietmed\_027\_a}     & \multicolumn{1}{c|}{\multirow{5}{*}{News}}   & \multicolumn{1}{c|}{\multirow{2}{*}{S00-T88}} & \multicolumn{1}{c|}{Host}          & \multicolumn{1}{c|}{F}             & SW            & 29              & 710             & 70.8                               & 20.8                              & 8.3                               & 6.2                               & 35.4                              & 100.0                               \\ \cline{1-1} \cline{4-14} 
\multicolumn{1}{|l|}{vietmed\_027\_b}     & \multicolumn{1}{c|}{}                        & \multicolumn{1}{c|}{}                         & \multicolumn{1}{c|}{Brc.}          & \multicolumn{1}{c|}{M}             & SE            & 64              & 1454            & 49.5                               & 39.1                              & 11.3                              & 5.6                               & 56.1                              & 100.0                               \\ \cline{1-1} \cline{3-14} 
\multicolumn{1}{|l|}{vietmed\_028\_a}     & \multicolumn{1}{c|}{}                        & \multicolumn{1}{c|}{\multirow{3}{*}{V00-Y99}} & \multicolumn{1}{c|}{Host}          & \multicolumn{1}{c|}{F}             & SE            & 106             & 2617            & 52.7                               & 34.7                              & 12.6                              & 4.1                               & 51.5                              & 100.0                               \\ \cline{1-1} \cline{4-14} 
\multicolumn{1}{|l|}{vietmed\_028\_b}     & \multicolumn{1}{c|}{}                        & \multicolumn{1}{c|}{}                         & \multicolumn{1}{c|}{Brc.}          & \multicolumn{1}{c|}{M}             & SE            & 21              & 475             & 47.6                               & 41.5                              & 10.9                              & 6.7                               & 59.2                              & 100.0                               \\ \cline{1-1} \cline{4-14} 
\multicolumn{1}{|l|}{vietmed\_029}        & \multicolumn{1}{c|}{}                        & \multicolumn{1}{c|}{}                         & \multicolumn{1}{c|}{Brc.}          & \multicolumn{1}{c|}{F}             & SE            & 92              & 2240            & 60.4                               & 30.0                              & 9.6                               & 5.4                               & 45.1                              & 100.0                               \\ \hline
\multicolumn{6}{|l|}{Sum/Avg}                                                                                                                                                                                                      & 3437            & 76136           & 54.2                               & 33.5                              & 12.3                              & 6.0                               & \textbf{51.8}                     & 99.9                                \\ \hline
\multicolumn{6}{|l|}{Mean}                                                                                                                                                                                                         & 127.3           & 2819.9          & 53.0                               & 33.2                              & 13.8                              & 6.4                               & 53.3                              & 100.0                               \\ \hline
\multicolumn{6}{|l|}{Standard Deviation}                                                                                                                                                                                           & 129.6           & 2743.3          & 11.4                               & 8.0                               & 5.2                               & 3.1                               & 12.1                              & 0.1                                 \\ \hline
\multicolumn{6}{|l|}{Median}                                                                                                                                                                                                       & 86.0            & 1838.0          & 55.3                               & 31.9                              & 12.8                              & 5.6                               & 51.5                              & 100.0                               \\ \hline
\end{tabular}%
}
\caption{\label{ASR_erroranalysis_XLSR53_test} Analysis of ASR errors on \textit{VietMed-Test} set using the baseline model \textit{XLSR-53} (WER = 51.8). 
\\Column from left to right is: Speaker ID, Recording Condition, ICD-10 Code, Speaker Role, Gender, Accent, Number of sentences, Number of words, Corrections, Substitution Errors, Deletion Errors, Insertion Errors, Word-Error-Rate, Sentence-Error-Rate. 
\\For Recording Condition, there are: Telephone (Tel.), Talkshow (Talk.), Podcast (Pod.), Diagnosis (Diag.), Lectures (Lec.), News.
\\For Speaker Role, there are: Lecturer (Lec.), Doctor (Doc.), Talkshow Host (Host), Patient (Pat.), Podcaster (Pod.), Broadcaster (Brc.).
\\For Gender, there are: Male (M) and Female (F).
\\For Accent, there are: South Central Coast (SCC), North (N), Southwest (SW), Southeast (SE), North Central Coast (NCC).}
\end{center}
\end{table*}
\newpage
\begin{table*}[h]
\begin{center}
\resizebox{\textwidth}{!}{%
\begin{tabular}{|lccccc|c|c|l|l|l|l|l|l|}
\hline
\multicolumn{1}{|c|}{\textbf{Speaker ID}} & \multicolumn{1}{c|}{\textbf{Rec.}}           & \multicolumn{1}{c|}{\textbf{ICD-10}}          & \multicolumn{1}{c|}{\textbf{Role}} & \multicolumn{1}{c|}{\textbf{Gend}} & \textbf{Acc.} & \textbf{\# Snt} & \textbf{\# Wrd} & \multicolumn{1}{c|}{\textbf{Corr}} & \multicolumn{1}{c|}{\textbf{Sub}} & \multicolumn{1}{c|}{\textbf{Del}} & \multicolumn{1}{c|}{\textbf{Ins}} & \multicolumn{1}{c|}{\textbf{Err}} & \multicolumn{1}{c|}{\textbf{S.Err}} \\ \hline
\multicolumn{1}{|l|}{vietmed\_002}        & \multicolumn{1}{c|}{\multirow{7}{*}{Tel.}}   & \multicolumn{1}{c|}{N00-N99}                  & \multicolumn{1}{c|}{Lec.}          & \multicolumn{1}{c|}{F}             & SCC           & 363             & 7631            & 33.8                               & 50.7                              & 15.5                              & 5.6                               & 71.8                              & 100.0                               \\ \cline{1-1} \cline{3-14} 
\multicolumn{1}{|l|}{vietmed\_004}        & \multicolumn{1}{c|}{}                        & \multicolumn{1}{c|}{M00-M99}                  & \multicolumn{1}{c|}{Doc.}          & \multicolumn{1}{c|}{M}             & SCC           & 446             & 10575           & 52.1                               & 34.1                              & 13.8                              & 6.5                               & 54.3                              & 100.0                               \\ \cline{1-1} \cline{3-14} 
\multicolumn{1}{|l|}{vietmed\_014\_a}     & \multicolumn{1}{c|}{}                        & \multicolumn{1}{c|}{\multirow{2}{*}{K00-K95}} & \multicolumn{1}{c|}{Host}          & \multicolumn{1}{c|}{F}             & N             & 18              & 491             & 72.3                               & 15.9                              & 11.8                              & 5.1                               & 32.8                              & 100.0                               \\ \cline{1-1} \cline{4-14} 
\multicolumn{1}{|l|}{vietmed\_014\_b}     & \multicolumn{1}{c|}{}                        & \multicolumn{1}{c|}{}                         & \multicolumn{1}{c|}{Doc.}          & \multicolumn{1}{c|}{M}             & N             & 164             & 4034            & 57.8                               & 28.6                              & 13.6                              & 4.6                               & 46.8                              & 100.0                               \\ \cline{1-1} \cline{3-14} 
\multicolumn{1}{|l|}{vietmed\_015\_a}     & \multicolumn{1}{c|}{}                        & \multicolumn{1}{c|}{\multirow{3}{*}{O00-O9A}} & \multicolumn{1}{c|}{Host}          & \multicolumn{1}{c|}{F}             & N             & 73              & 1779            & 70.8                               & 18.1                              & 11.1                              & 4.5                               & 33.7                              & 100.0                               \\ \cline{1-1} \cline{4-14} 
\multicolumn{1}{|l|}{vietmed\_015\_b}     & \multicolumn{1}{c|}{}                        & \multicolumn{1}{c|}{}                         & \multicolumn{1}{c|}{Doc.}          & \multicolumn{1}{c|}{F}             & N             & 297             & 5669            & 60.1                               & 26.7                              & 13.2                              & 4.7                               & 44.6                              & 99.7                                \\ \cline{1-1} \cline{4-14} 
\multicolumn{1}{|l|}{vietmed\_015\_c}     & \multicolumn{1}{c|}{}                        & \multicolumn{1}{c|}{}                         & \multicolumn{1}{c|}{Pat.}          & \multicolumn{1}{c|}{F}             & N             & 55              & 1010            & 44.4                               & 37.5                              & 18.1                              & 5.4                               & 61.1                              & 100.0                               \\ \hline
\multicolumn{1}{|l|}{vietmed\_017\_a}     & \multicolumn{1}{c|}{\multirow{10}{*}{Talk.}} & \multicolumn{1}{c|}{\multirow{2}{*}{U00-U85}} & \multicolumn{1}{c|}{Doc.}          & \multicolumn{1}{c|}{F}             & SW            & 47              & 1104            & 51.6                               & 36.2                              & 12.1                              & 6.3                               & 54.6                              & 100.0                               \\ \cline{1-1} \cline{4-14} 
\multicolumn{1}{|l|}{vietmed\_017\_b}     & \multicolumn{1}{c|}{}                        & \multicolumn{1}{c|}{}                         & \multicolumn{1}{c|}{Doc.}          & \multicolumn{1}{c|}{M}             & N             & 86              & 2061            & 62.4                               & 26.7                              & 10.9                              & 4.9                               & 42.4                              & 100.0                               \\ \cline{1-1} \cline{3-14} 
\multicolumn{1}{|l|}{vietmed\_018\_a}     & \multicolumn{1}{c|}{}                        & \multicolumn{1}{c|}{\multirow{6}{*}{K00-K95}} & \multicolumn{1}{c|}{Host}          & \multicolumn{1}{c|}{F}             & SW            & 63              & 1527            & 59.2                               & 27.5                              & 13.3                              & 19.6                              & 60.4                              & 100.0                               \\ \cline{1-1} \cline{4-14} 
\multicolumn{1}{|l|}{vietmed\_018\_b}     & \multicolumn{1}{c|}{}                        & \multicolumn{1}{c|}{}                         & \multicolumn{1}{c|}{Doc.}          & \multicolumn{1}{c|}{M}             & SW            & 192             & 5293            & 59.5                               & 26.3                              & 14.3                              & 6.7                               & 47.2                              & 100.0                               \\ \cline{1-1} \cline{4-14} 
\multicolumn{1}{|l|}{vietmed\_018\_c}     & \multicolumn{1}{c|}{}                        & \multicolumn{1}{c|}{}                         & \multicolumn{1}{c|}{Doc.}          & \multicolumn{1}{c|}{F}             & SW            & 118             & 2761            & 57.7                               & 29.6                              & 12.7                              & 9.0                               & 51.4                              & 100.0                               \\ \cline{1-1} \cline{4-14} 
\multicolumn{1}{|l|}{vietmed\_018\_d}     & \multicolumn{1}{c|}{}                        & \multicolumn{1}{c|}{}                         & \multicolumn{1}{c|}{Pat.}          & \multicolumn{1}{c|}{F}             & SW            & 20              & 412             & 34.7                               & 34.5                              & 30.8                              & 4.9                               & 70.1                              & 100.0                               \\ \cline{1-1} \cline{4-14} 
\multicolumn{1}{|l|}{vietmed\_018\_e}     & \multicolumn{1}{c|}{}                        & \multicolumn{1}{c|}{}                         & \multicolumn{1}{c|}{Pat.}          & \multicolumn{1}{c|}{M}             & SW            & 5               & 76              & 42.1                               & 34.2                              & 23.7                              & 7.9                               & 65.8                              & 100.0                               \\ \cline{1-1} \cline{4-14} 
\multicolumn{1}{|l|}{vietmed\_018\_f}     & \multicolumn{1}{c|}{}                        & \multicolumn{1}{c|}{}                         & \multicolumn{1}{c|}{Doc.}          & \multicolumn{1}{c|}{M}             & SW            & 25              & 639             & 44.0                               & 38.2                              & 17.8                              & 7.0                               & 63.1                              & 100.0                               \\ \cline{1-1} \cline{3-14} 
\multicolumn{1}{|l|}{vietmed\_019\_a}     & \multicolumn{1}{c|}{}                        & \multicolumn{1}{c|}{\multirow{2}{*}{L00-L99}} & \multicolumn{1}{c|}{Host}          & \multicolumn{1}{c|}{F}             & SW            & 58              & 1490            & 58.6                               & 28.7                              & 12.7                              & 6.8                               & 48.2                              & 100.0                               \\ \cline{1-1} \cline{4-14} 
\multicolumn{1}{|l|}{vietmed\_019\_b}     & \multicolumn{1}{c|}{}                        & \multicolumn{1}{c|}{}                         & \multicolumn{1}{c|}{Doc.}          & \multicolumn{1}{c|}{F}             & SW            & 116             & 2776            & 58.9                               & 28.4                              & 12.7                              & 7.4                               & 48.5                              & 100.0                               \\ \hline
\multicolumn{1}{|l|}{vietmed\_023}        & \multicolumn{1}{c|}{\multirow{2}{*}{Pod.}}   & \multicolumn{1}{c|}{P00-P96}                  & \multicolumn{1}{c|}{Pod.}          & \multicolumn{1}{c|}{F}             & SW            & 390             & 7414            & 63.0                               & 29.6                              & 7.4                               & 4.8                               & 41.8                              & 99.7                                \\ \cline{1-1} \cline{3-14} 
\multicolumn{1}{|l|}{vietmed\_024}        & \multicolumn{1}{c|}{}                        & \multicolumn{1}{c|}{O00\_O99}                 & \multicolumn{1}{c|}{Pod.}          & \multicolumn{1}{c|}{F}             & SE            & 376             & 7425            & 65.4                               & 25.9                              & 8.6                               & 5.8                               & 40.3                              & 99.5                                \\ \hline
\multicolumn{1}{|l|}{vietmed\_025\_a}     & \multicolumn{1}{c|}{\multirow{2}{*}{Diag.}}  & \multicolumn{1}{c|}{\multirow{2}{*}{H60-H95}} & \multicolumn{1}{c|}{Host}          & \multicolumn{1}{c|}{F}             & SW            & 101             & 2280            & 65.3                               & 24.5                              & 10.2                              & 4.6                               & 39.3                              & 100.0                               \\ \cline{1-1} \cline{4-14} 
\multicolumn{1}{|l|}{vietmed\_025\_b}     & \multicolumn{1}{c|}{}                        & \multicolumn{1}{c|}{}                         & \multicolumn{1}{c|}{Doc.}          & \multicolumn{1}{c|}{M}             & SE            & 91              & 1838            & 67.2                               & 23.2                              & 9.5                               & 7.1                               & 39.8                              & 100.0                               \\ \hline
\multicolumn{1}{|l|}{vietmed\_026}        & \multicolumn{1}{c|}{Lec.}                    & \multicolumn{1}{c|}{A00-B99}                  & \multicolumn{1}{c|}{Lec.}          & \multicolumn{1}{c|}{M}             & NCC           & 21              & 355             & 26.5                               & 47.3                              & 26.2                              & 4.8                               & 78.3                              & 100.0                               \\ \hline
\multicolumn{1}{|l|}{vietmed\_027\_a}     & \multicolumn{1}{c|}{\multirow{5}{*}{News}}   & \multicolumn{1}{c|}{\multirow{2}{*}{S00-T88}} & \multicolumn{1}{c|}{Host}          & \multicolumn{1}{c|}{F}             & SW            & 29              & 710             & 68.7                               & 22.5                              & 8.7                               & 5.5                               & 36.8                              & 100.0                               \\ \cline{1-1} \cline{4-14} 
\multicolumn{1}{|l|}{vietmed\_027\_b}     & \multicolumn{1}{c|}{}                        & \multicolumn{1}{c|}{}                         & \multicolumn{1}{c|}{Brc.}          & \multicolumn{1}{c|}{M}             & SE            & 64              & 1454            & 41.5                               & 44.6                              & 13.9                              & 5.2                               & 63.7                              & 100.0                               \\ \cline{1-1} \cline{3-14} 
\multicolumn{1}{|l|}{vietmed\_028\_a}     & \multicolumn{1}{c|}{}                        & \multicolumn{1}{c|}{\multirow{3}{*}{V00-Y99}} & \multicolumn{1}{c|}{Host}          & \multicolumn{1}{c|}{F}             & SE            & 106             & 2617            & 59.7                               & 28.8                              & 11.5                              & 4.4                               & 44.7                              & 99.1                                \\ \cline{1-1} \cline{4-14} 
\multicolumn{1}{|l|}{vietmed\_028\_b}     & \multicolumn{1}{c|}{}                        & \multicolumn{1}{c|}{}                         & \multicolumn{1}{c|}{Brc.}          & \multicolumn{1}{c|}{M}             & SE            & 21              & 475             & 48.8                               & 39.2                              & 12.0                              & 5.1                               & 56.2                              & 100.0                               \\ \cline{1-1} \cline{4-14} 
\multicolumn{1}{|l|}{vietmed\_029}        & \multicolumn{1}{c|}{}                        & \multicolumn{1}{c|}{}                         & \multicolumn{1}{c|}{Brc.}          & \multicolumn{1}{c|}{F}             & SE            & 92              & 2240            & 64.4                               & 26.1                              & 9.6                               & 5.9                               & 41.6                              & 100.0                               \\ \hline
\multicolumn{6}{|l|}{Sum/Avg}                                                                                                                                                                                                      & 3437            & 76136           & 56.5                               & 31.2                              & 12.3                              & 6.0                               & \textbf{49.5}                     & 99.9                                \\ \hline
\multicolumn{6}{|l|}{Mean}                                                                                                                                                                                                         & 127.3           & 2819.9          & 55.2                               & 30.9                              & 13.9                              & 6.3                               & 51.1                              & 99.9                                \\ \hline
\multicolumn{6}{|l|}{Standard Deviation}                                                                                                                                                                                           & 129.6           & 2743.3          & 12.0                               & 8.3                               & 5.4                               & 2.9                               & 12.2                              & 0.2                                 \\ \hline
\multicolumn{6}{|l|}{Median}                                                                                                                                                                                                       & 86.0            & 1838.0          & 58.9                               & 28.7                              & 12.7                              & 5.5                               & 48.2                              & 100.0                               \\ \hline
\end{tabular}%
}
\caption{\label{ASR_erroranalysis_w2v2Viet_test} Analysis of ASR errors on \textit{VietMed-Test} set using the baseline model \textit{w2v2-Viet} (WER = 49.5). 
\\Column from left to right is: Speaker ID, Recording Condition, ICD-10 Code, Speaker Role, Gender, Accent, Number of sentences, Number of words, Corrections, Substitution Errors, Deletion Errors, Insertion Errors, Word-Error-Rate, Sentence-Error-Rate. 
\\For Recording Condition, there are: Telephone (Tel.), Talkshow (Talk.), Podcast (Pod.), Diagnosis (Diag.), Lectures (Lec.), News.
\\For Speaker Role, there are: Lecturer (Lec.), Doctor (Doc.), Talkshow Host (Host), Patient (Pat.), Podcaster (Pod.), Broadcaster (Brc.).
\\For Gender, there are: Male (M) and Female (F).
\\For Accent, there are: South Central Coast (SCC), North (N), Southwest (SW), Southeast (SE), North Central Coast (NCC).}
\end{center}
\end{table*}
\newpage
\begin{table*}[h]
\begin{center}
\resizebox{\textwidth}{!}{%
\begin{tabular}{|lccccc|c|c|c|c|c|c|c|c|}
\hline
\multicolumn{1}{|c|}{\textbf{Speaker ID}} & \multicolumn{1}{c|}{\textbf{Rec.}}           & \multicolumn{1}{c|}{\textbf{ICD-10}}          & \multicolumn{1}{c|}{\textbf{Role}} & \multicolumn{1}{c|}{\textbf{Gend}} & \textbf{Acc.} & \textbf{\# Snt} & \textbf{\# Wrd} & \textbf{Corr} & \textbf{Sub} & \textbf{Del} & \textbf{Ins} & \textbf{Err} & \textbf{S.Err} \\ \hline
\multicolumn{1}{|l|}{vietmed\_002}        & \multicolumn{1}{c|}{\multirow{7}{*}{Tel.}}   & \multicolumn{1}{c|}{N00-N99}                  & \multicolumn{1}{c|}{Lec.}          & \multicolumn{1}{c|}{F}              & SCC             & 363             & 7631            & 57.8          & 31.2         & 11.0         & 6.3          & 48.5         & 100.0          \\ \cline{1-1} \cline{3-14} 
\multicolumn{1}{|l|}{vietmed\_004}        & \multicolumn{1}{c|}{}                        & \multicolumn{1}{c|}{M00-M99}                  & \multicolumn{1}{c|}{Doc.}          & \multicolumn{1}{c|}{M}              & SCC             & 446             & 10575           & 68.8          & 18.7         & 12.5         & 5.4          & 36.6         & 100.0          \\ \cline{1-1} \cline{3-14} 
\multicolumn{1}{|l|}{vietmed\_014\_a}     & \multicolumn{1}{c|}{}                        & \multicolumn{1}{c|}{\multirow{2}{*}{K00-K95}} & \multicolumn{1}{c|}{Host}          & \multicolumn{1}{c|}{F}              & N               & 18              & 491             & 87.8          & 3.5          & 8.8          & 4.7          & 16.9         & 100.0          \\ \cline{1-1} \cline{4-14} 
\multicolumn{1}{|l|}{vietmed\_014\_b}     & \multicolumn{1}{c|}{}                        & \multicolumn{1}{c|}{}                         & \multicolumn{1}{c|}{Doc.}          & \multicolumn{1}{c|}{M}              & N               & 164             & 4034            & 77.2          & 12.2         & 10.5         & 4.6          & 27.4         & 100.0          \\ \cline{1-1} \cline{3-14} 
\multicolumn{1}{|l|}{vietmed\_015\_a}     & \multicolumn{1}{c|}{}                        & \multicolumn{1}{c|}{\multirow{3}{*}{O00-O9A}} & \multicolumn{1}{c|}{Host}          & \multicolumn{1}{c|}{F}              & N               & 73              & 1779            & 85.2          & 5.8          & 9.0          & 3.6          & 18.4         & 97.3           \\ \cline{1-1} \cline{4-14} 
\multicolumn{1}{|l|}{vietmed\_015\_b}     & \multicolumn{1}{c|}{}                        & \multicolumn{1}{c|}{}                         & \multicolumn{1}{c|}{Doc.}          & \multicolumn{1}{c|}{F}              & N               & 297             & 5669            & 82.4          & 7.7          & 9.8          & 4.2          & 21.8         & 97.3           \\ \cline{1-1} \cline{4-14} 
\multicolumn{1}{|l|}{vietmed\_015\_c}     & \multicolumn{1}{c|}{}                        & \multicolumn{1}{c|}{}                         & \multicolumn{1}{c|}{Pat.}          & \multicolumn{1}{c|}{F}              & N               & 55              & 1010            & 70.1          & 14.9         & 15.0         & 5.8          & 35.7         & 100.0          \\ \hline
\multicolumn{1}{|l|}{vietmed\_017\_a}     & \multicolumn{1}{c|}{\multirow{10}{*}{Talk.}} & \multicolumn{1}{c|}{\multirow{2}{*}{U00-U85}} & \multicolumn{1}{c|}{Doc.}          & \multicolumn{1}{c|}{F}              & SW              & 47              & 1104            & 76.6          & 13.1         & 10.2         & 4.2          & 27.5         & 100.0          \\ \cline{1-1} \cline{4-14} 
\multicolumn{1}{|l|}{vietmed\_017\_b}     & \multicolumn{1}{c|}{}                        & \multicolumn{1}{c|}{}                         & \multicolumn{1}{c|}{Doc.}          & \multicolumn{1}{c|}{M}              & N               & 86              & 2061            & 80.1          & 10.4         & 9.6          & 4.8          & 24.7         & 100.0          \\ \cline{1-1} \cline{3-14} 
\multicolumn{1}{|l|}{vietmed\_018\_a}     & \multicolumn{1}{c|}{}                        & \multicolumn{1}{c|}{\multirow{6}{*}{K00-K95}} & \multicolumn{1}{c|}{Host}          & \multicolumn{1}{c|}{F}              & SW              & 63              & 1527            & 73.7          & 13.2         & 13.2         & 18.7         & 45.1         & 100.0          \\ \cline{1-1} \cline{4-14} 
\multicolumn{1}{|l|}{vietmed\_018\_b}     & \multicolumn{1}{c|}{}                        & \multicolumn{1}{c|}{}                         & \multicolumn{1}{c|}{Doc.}          & \multicolumn{1}{c|}{M}              & SW              & 192             & 5293            & 75.3          & 12.1         & 12.6         & 6.5          & 31.2         & 100.0          \\ \cline{1-1} \cline{4-14} 
\multicolumn{1}{|l|}{vietmed\_018\_c}     & \multicolumn{1}{c|}{}                        & \multicolumn{1}{c|}{}                         & \multicolumn{1}{c|}{Doc.}          & \multicolumn{1}{c|}{F}              & SW              & 118             & 2761            & 74.3          & 12.4         & 13.3         & 7.3          & 33.0         & 100.0          \\ \cline{1-1} \cline{4-14} 
\multicolumn{1}{|l|}{vietmed\_018\_d}     & \multicolumn{1}{c|}{}                        & \multicolumn{1}{c|}{}                         & \multicolumn{1}{c|}{Pat.}          & \multicolumn{1}{c|}{F}              & SW              & 20              & 412             & 55.1          & 20.6         & 24.3         & 5.6          & 50.5         & 100.0          \\ \cline{1-1} \cline{4-14} 
\multicolumn{1}{|l|}{vietmed\_018\_e}     & \multicolumn{1}{c|}{}                        & \multicolumn{1}{c|}{}                         & \multicolumn{1}{c|}{Pat.}          & \multicolumn{1}{c|}{M}              & SW              & 5               & 76              & 57.9          & 19.7         & 22.4         & 7.9          & 50.0         & 100.0          \\ \cline{1-1} \cline{4-14} 
\multicolumn{1}{|l|}{vietmed\_018\_f}     & \multicolumn{1}{c|}{}                        & \multicolumn{1}{c|}{}                         & \multicolumn{1}{c|}{Doc.}          & \multicolumn{1}{c|}{M}              & SW              & 25              & 639             & 64.9          & 20.3         & 14.7         & 6.1          & 41.2         & 100.0          \\ \cline{1-1} \cline{3-14} 
\multicolumn{1}{|l|}{vietmed\_019\_a}     & \multicolumn{1}{c|}{}                        & \multicolumn{1}{c|}{\multirow{2}{*}{L00-L99}} & \multicolumn{1}{c|}{Host}          & \multicolumn{1}{c|}{F}              & SW              & 58              & 1490            & 75.2          & 12.6         & 12.2         & 6.7          & 31.5         & 100.0          \\ \cline{1-1} \cline{4-14} 
\multicolumn{1}{|l|}{vietmed\_019\_b}     & \multicolumn{1}{c|}{}                        & \multicolumn{1}{c|}{}                         & \multicolumn{1}{c|}{Doc.}          & \multicolumn{1}{c|}{F}              & SW              & 116             & 2776            & 75.7          & 11.9         & 12.5         & 6.2          & 30.5         & 100.0          \\ \hline
\multicolumn{1}{|l|}{vietmed\_023}        & \multicolumn{1}{c|}{\multirow{2}{*}{Pod.}}   & \multicolumn{1}{c|}{P00-P96}                  & \multicolumn{1}{c|}{Pod.}          & \multicolumn{1}{c|}{F}              & SW              & 390             & 7414            & 83.3          & 10.6         & 6.0          & 4.1          & 20.8         & 97.4           \\ \cline{1-1} \cline{3-14} 
\multicolumn{1}{|l|}{vietmed\_024}        & \multicolumn{1}{c|}{}                        & \multicolumn{1}{c|}{O00\_O99}                 & \multicolumn{1}{c|}{Pod.}          & \multicolumn{1}{c|}{F}              & SE              & 376             & 7425            & 85.0          & 8.0          & 7.1          & 5.0          & 20.1         & 98.4           \\ \hline
\multicolumn{1}{|l|}{vietmed\_025\_a}     & \multicolumn{1}{c|}{\multirow{2}{*}{Diag.}}  & \multicolumn{1}{c|}{\multirow{2}{*}{H60-H95}} & \multicolumn{1}{c|}{Host}          & \multicolumn{1}{c|}{F}              & SW              & 101             & 2280            & 80.4          & 10.6         & 9.0          & 4.8          & 24.4         & 100.0          \\ \cline{1-1} \cline{4-14} 
\multicolumn{1}{|l|}{vietmed\_025\_b}     & \multicolumn{1}{c|}{}                        & \multicolumn{1}{c|}{}                         & \multicolumn{1}{c|}{Doc.}          & \multicolumn{1}{c|}{M}              & SE              & 91              & 1838            & 81.8          & 10.0         & 8.3          & 5.1          & 23.3         & 98.9           \\ \hline
\multicolumn{1}{|l|}{vietmed\_026}        & \multicolumn{1}{c|}{Lec.}                    & \multicolumn{1}{c|}{A00-B99}                  & \multicolumn{1}{c|}{Lec.}          & \multicolumn{1}{c|}{M}              & NCC             & 21              & 355             & 57.7          & 27.9         & 14.4         & 7.3          & 49.6         & 100.0          \\ \hline
\multicolumn{1}{|l|}{vietmed\_027\_a}     & \multicolumn{1}{c|}{\multirow{5}{*}{News}}   & \multicolumn{1}{c|}{\multirow{2}{*}{S00-T88}} & \multicolumn{1}{c|}{Host}          & \multicolumn{1}{c|}{F}              & SW              & 29              & 710             & 83.5          & 8.0          & 8.5          & 4.6          & 21.1         & 100.0          \\ \cline{1-1} \cline{4-14} 
\multicolumn{1}{|l|}{vietmed\_027\_b}     & \multicolumn{1}{c|}{}                        & \multicolumn{1}{c|}{}                         & \multicolumn{1}{c|}{Brc.}          & \multicolumn{1}{c|}{M}              & SE              & 64              & 1454            & 74.8          & 15.8         & 9.4          & 5.2          & 30.4         & 100.0          \\ \cline{1-1} \cline{3-14} 
\multicolumn{1}{|l|}{vietmed\_028\_a}     & \multicolumn{1}{c|}{}                        & \multicolumn{1}{c|}{\multirow{3}{*}{V00-Y99}} & \multicolumn{1}{c|}{Host}          & \multicolumn{1}{c|}{F}              & SE              & 106             & 2617            & 82.7          & 8.8          & 8.6          & 4.2          & 21.6         & 99.1           \\ \cline{1-1} \cline{4-14} 
\multicolumn{1}{|l|}{vietmed\_028\_b}     & \multicolumn{1}{c|}{}                        & \multicolumn{1}{c|}{}                         & \multicolumn{1}{c|}{Brc.}          & \multicolumn{1}{c|}{M}              & SE              & 21              & 475             & 74.3          & 14.9         & 10.7         & 5.3          & 30.9         & 100.0          \\ \cline{1-1} \cline{4-14} 
\multicolumn{1}{|l|}{vietmed\_029}        & \multicolumn{1}{c|}{}                        & \multicolumn{1}{c|}{}                         & \multicolumn{1}{c|}{Brc.}          & \multicolumn{1}{c|}{F}              & SE              & 92              & 2240            & 83.9          & 7.5          & 8.5          & 5.6          & 21.7         & 97.8           \\ \hline
\multicolumn{6}{|l|}{Sum/Avg}                                                                                                                                                                                                         & 3437            & 76136           & 75.9          & 13.8         & 10.3         & 5.6          & \textbf{29.6}         & 99.1           \\ \hline
\multicolumn{6}{|l|}{Mean}                                                                                                                                                                                                            & 127.3           & 2819.9          & 75.0          & 13.4         & 11.6         & 5.9          & 30.9         & 99.5           \\ \hline
\multicolumn{6}{|l|}{Standard Deviation}                                                                                                                                                                                              & 129.6           & 2743.3          & 9.3           & 6.4          & 4.1          & 2.8          & 10.5         & 1.0            \\ \hline
\multicolumn{6}{|l|}{Median}                                                                                                                                                                                                          & 86.0            & 1838.0          & 75.7          & 12.2         & 10.5         & 5.3          & 30.4         & 100.0          \\ \hline
\end{tabular}%
}
\caption{\label{ASR_erroranalysis_XLSR53Viet_test} Analysis of ASR errors on \textit{VietMed-Test} set using the best baseline model \textit{XLSR-53-Viet} (WER = 29.6). 
\\Column from left to right is: Speaker ID, Recording Condition, ICD-10 Code, Speaker Role, Gender, Accent, Number of sentences, Number of words, Corrections, Substitution Errors, Deletion Errors, Insertion Errors, Word-Error-Rate, Sentence-Error-Rate. 
\\For Recording Condition, there are: Telephone (Tel.), Talkshow (Talk.), Podcast (Pod.), Diagnosis (Diag.), Lectures (Lec.), News.
\\For Speaker Role, there are: Lecturer (Lec.), Doctor (Doc.), Talkshow Host (Host), Patient (Pat.), Podcaster (Pod.), Broadcaster (Brc.).
\\For Gender, there are: Male (M) and Female (F).
\\For Accent, there are: South Central Coast (SCC), North (N), Southwest (SW), Southeast (SE), North Central Coast (NCC).}
\end{center}
\end{table*}
\newpage

\begin{longtable}{|c|c|l|c|}
\hline
\textbf{Index} & \textbf{Occurrences} & \multicolumn{1}{c|}{\textbf{Confusion pair}} & \textbf{Type} \\ \hline
\endfirsthead
\multicolumn{4}{c}%
{{\bfseries Table \thetable\ continued from previous page}} \\
\hline
\textbf{Index} & \textbf{Occurrences} & \multicolumn{1}{c|}{\textbf{Confusion pair}} & \textbf{Type} \\ \hline
\endhead
1   & 75 & bé $\Longrightarrow$ béo         & Med  \\ \hline
2   & 75 & cung $\Longrightarrow$ công      &  -    \\ \hline
3   & 49 & các $\Longrightarrow$ cái        &  -    \\ \hline
4   & 34 & trẻ $\Longrightarrow$ sẽ         & Med  \\ \hline
5   & 33 & bú $\Longrightarrow$ bốn         & Med  \\ \hline
6   & 31 & implant $\Longrightarrow$ lên    & Med  \\ \hline
7   & 30 & thai $\Longrightarrow$ hai       & Med  \\ \hline
8   & 28 & cái $\Longrightarrow$ các        & Fill \\ \hline
9   & 26 & là $\Longrightarrow$ mà          & Fill \\ \hline
10  & 25 & tử $\Longrightarrow$ bệnh        & Med  \\ \hline
11  & 25 & vì $\Longrightarrow$ thì         & Fill \\ \hline
12  & 24 & răng $\Longrightarrow$ đang      & Med  \\ \hline
13  & 23 & cấy $\Longrightarrow$ cái        & Med  \\ \hline
14  & 23 & làm $\Longrightarrow$ là         &  -    \\ \hline
15  & 21 & là $\Longrightarrow$ và          & Fill \\ \hline
16  & 20 & đó $\Longrightarrow$ nó          & Fill \\ \hline
17  & 19 & và $\Longrightarrow$ là          & Fill \\ \hline
18  & 19 & và $\Longrightarrow$ mà          & Fill \\ \hline
19  & 19 & âm $\Longrightarrow$ ăn          & Med  \\ \hline
20  & 18 & là $\Longrightarrow$ làm         & Fill \\ \hline
21  & 18 & mình $\Longrightarrow$ mà        & Fill \\ \hline
22  & 18 & trồng $\Longrightarrow$ trong    & Med  \\ \hline
23  & 17 & bú $\Longrightarrow$ bố          & Med  \\ \hline
24  & 17 & chị $\Longrightarrow$ chỉ        &  -    \\ \hline
25  & 17 & có $\Longrightarrow$ cái         &  -    \\ \hline
26  & 17 & là $\Longrightarrow$ lại         & Fill \\ \hline
27  & 17 & mà $\Longrightarrow$ và          & Fill \\ \hline
28  & 17 & sẽ $\Longrightarrow$ phải        & Fill \\ \hline
29  & 17 & đi $\Longrightarrow$ đây         & Fill \\ \hline
30  & 16 & nó $\Longrightarrow$ đó          & Fill \\ \hline
31  & 16 & tử $\Longrightarrow$ về          & Med  \\ \hline
32  & 15 & con $\Longrightarrow$ còn        & Med  \\ \hline
33            & 15                                 & progesterone $\Longrightarrow$ cholesterol      & Med           \\ \hline
34  & 15 & rong $\Longrightarrow$ năm       & Med  \\ \hline
35  & 15 & thủ $\Longrightarrow$ phẫu       & Med  \\ \hline
36  & 14 & implant $\Longrightarrow$ selen  & Med  \\ \hline
37  & 14 & que $\Longrightarrow$ quen       & Med  \\ \hline
38  & 13 & còn $\Longrightarrow$ có         & Fill \\ \hline
39  & 13 & có $\Longrightarrow$ các         & Fill \\ \hline
40  & 13 & có $\Longrightarrow$ đó          & Fill \\ \hline
41  & 13 & lại $\Longrightarrow$ là         & Fill \\ \hline
42  & 12 & như $\Longrightarrow$ nhưng      & Fill \\ \hline
43  & 11 & bà $\Longrightarrow$ mà          & -     \\ \hline
44  & 11 & bình $\Longrightarrow$ bệnh      & Med  \\ \hline
45  & 11 & cung $\Longrightarrow$ trong     & Med  \\ \hline
46  & 11 & là $\Longrightarrow$ nó          & Fill \\ \hline
47  & 11 & mình $\Longrightarrow$ bệnh      &  -    \\ \hline
48  & 11 & răng $\Longrightarrow$ gan       & Med  \\ \hline
49  & 11 & răng $\Longrightarrow$ ăn        & Med  \\ \hline
50  & 11 & vào $\Longrightarrow$ và         &  -    \\ \hline
51  & 10 & anh $\Longrightarrow$ ăn         &  -    \\ \hline
52  & 10 & bà $\Longrightarrow$ ba          &  -    \\ \hline
53  & 10 & chú $\Longrightarrow$ chúng      &   -   \\ \hline
54  & 10 & cách $\Longrightarrow$ các       &  -    \\ \hline
55  & 10 & cô $\Longrightarrow$ của         &  -    \\ \hline
56  & 10 & da $\Longrightarrow$ ra          & Med  \\ \hline
57  & 10 & khi $\Longrightarrow$ thì        & -     \\ \hline
58  & 10 & lạ $\Longrightarrow$ là          &  -    \\ \hline
59  & 10 & tóc $\Longrightarrow$ tác        & Med  \\ \hline
60  & 10 & vòng $\Longrightarrow$ phòng     &  -    \\ \hline
61  & 10 & đo $\Longrightarrow$ đó          & Med  \\ \hline
62  & 10 & đại $\Longrightarrow$ tại        &  -    \\ \hline
63  & 9  & cổ $\Longrightarrow$ của         & Med  \\ \hline
64  & 9  & dặm $\Longrightarrow$ giảm       & Med  \\ \hline
65  & 9  & hay $\Longrightarrow$ hai        &  -    \\ \hline
66  & 9  & ngừa $\Longrightarrow$ là        & Med  \\ \hline
67  & 9  & nói $\Longrightarrow$ nó         &  -    \\ \hline
68  & 9  & răng $\Longrightarrow$ rằng      & Med  \\ \hline
69  & 9  & sau $\Longrightarrow$ sao        & -     \\ \hline
70  & 9  & tai $\Longrightarrow$ tay        & Med  \\ \hline
71  & 9  & thì $\Longrightarrow$ cái        & Fill \\ \hline
72  & 9  & tràng $\Longrightarrow$ trạm     & Med  \\ \hline
73  & 9  & tóc $\Longrightarrow$ tắt        & Med  \\ \hline
74  & 9  & ốc $\Longrightarrow$ cái         & Med  \\ \hline
75  & 8  & chị $\Longrightarrow$ thì        &  -    \\ \hline
76  & 8  & cong $\Longrightarrow$ công      & Med  \\ \hline
77  & 8  & em $\Longrightarrow$ xem         & -     \\ \hline
78  & 8  & estrogen $\Longrightarrow$ selen & Med  \\ \hline
79  & 8  & kinh $\Longrightarrow$ cân       & Med  \\ \hline
80  & 8  & nhi $\Longrightarrow$ như        & Med  \\ \hline
81  & 8  & nè $\Longrightarrow$ này         & Fill \\ \hline
82  & 8  & quy $\Longrightarrow$ quá        & Med  \\ \hline
83  & 8  & ruột $\Longrightarrow$ rồi       & Med  \\ \hline
84  & 8  & răng $\Longrightarrow$ năng      & Med  \\ \hline
85  & 8  & tai $\Longrightarrow$ ta         & Med  \\ \hline
86  & 8  & thật $\Longrightarrow$ thực      & -     \\ \hline
87  & 8  & thể $\Longrightarrow$ thế        & Med  \\ \hline
88  & 8  & trồng $\Longrightarrow$ chọn     & Med  \\ \hline
89  & 8  & tóc $\Longrightarrow$ tốt        & Med  \\ \hline
90  & 8  & tự $\Longrightarrow$ từ          & Med  \\ \hline
91  & 8  & và $\Longrightarrow$ vào         & Fill \\ \hline
92  & 8  & để $\Longrightarrow$ đến         & Fill \\ \hline
93  & 7  & an $\Longrightarrow$ ăn          &  -    \\ \hline
94  & 7  & bạn $\Longrightarrow$ bệnh       & -     \\ \hline
95  & 7  & canxi $\Longrightarrow$ xây      & Med  \\ \hline
96  & 7  & cho $\Longrightarrow$ cái        &  -    \\ \hline
97  & 7  & cái $\Longrightarrow$ có         & Fill \\ \hline
98  & 7  & có $\Longrightarrow$ tốt         & Fill \\ \hline
99  & 7  & cơn $\Longrightarrow$ cân        & Med  \\ \hline
100 & 7  & dày $\Longrightarrow$ dài        & Med  \\ \hline
101 & 7  & ghép $\Longrightarrow$ kết       & Med  \\ \hline
102 & 7  & già $\Longrightarrow$ ra         & Med  \\ \hline
103 & 7  & kinh $\Longrightarrow$ đến       & Med  \\ \hline
104 & 7  & kỹ $\Longrightarrow$ cái         &  -    \\ \hline
105 & 7  & là $\Longrightarrow$ ta          & Fill \\ \hline
106 & 7  & nữ $\Longrightarrow$ nữa         &  -    \\ \hline
107 & 7  & qua $\Longrightarrow$ quá        &  -    \\ \hline
108 & 7  & siêu $\Longrightarrow$ thức      & Med  \\ \hline
109 & 7  & thì $\Longrightarrow$ vì         & Fill \\ \hline
110 & 7  & thì $\Longrightarrow$ để         & Fill \\ \hline
111 & 7  & tử $\Longrightarrow$ thành       & Med  \\ \hline
112 & 7  & vậy $\Longrightarrow$ mà         & Fill \\ \hline
113 & 7  & vắcxin $\Longrightarrow$ sĩ      & Med  \\ \hline
114 & 7  & âm $\Longrightarrow$ tâm         & Med  \\ \hline
115 & 7  & đó $\Longrightarrow$ nữa         & Fill \\ \hline
116 & 7  & để $\Longrightarrow$ cái         & Fill \\ \hline
117 & 6  & buồng $\Longrightarrow$ buồn     & Med  \\ \hline
118 & 6  & bà $\Longrightarrow$ và          & -     \\ \hline
119 & 6  & cho $\Longrightarrow$ chất       &  -    \\ \hline
120 & 6  & cho $\Longrightarrow$ ra         &  -    \\ \hline
121 & 6  & con $\Longrightarrow$ có         & Med  \\ \hline
122 & 6  & cung $\Longrightarrow$ không     & Med  \\ \hline
123 & 6  & cách $\Longrightarrow$ cái       &  -    \\ \hline
124 & 6  & cái $\Longrightarrow$ với        & Fill \\ \hline
125 & 6  & có $\Longrightarrow$ của         & Fill \\ \hline
126 & 6  & có $\Longrightarrow$ nó          & -     \\ \hline
127 & 6  & cấy $\Longrightarrow$ thấy       & Med  \\ \hline
128 & 6  & của $\Longrightarrow$ có         & -     \\ \hline
129 & 6  & d $\Longrightarrow$ b            &  -    \\ \hline
130 & 6  & dịch $\Longrightarrow$ việc      & Med  \\ \hline
131 & 6  & f0 $\Longrightarrow$ không       & Med  \\ \hline
132 & 6  & ghép $\Longrightarrow$ biết      & Med  \\ \hline
133 & 6  & hợp $\Longrightarrow$ hai        & -     \\ \hline
134 & 6  & khiếm $\Longrightarrow$ khiến    &  -    \\ \hline
135 & 6  & khá $\Longrightarrow$ khác       &  -    \\ \hline
136 & 6  & lý $\Longrightarrow$ lấy         &  -    \\ \hline
137 & 6  & lạ $\Longrightarrow$ lại         &   -   \\ \hline
138 & 6  & mãn $\Longrightarrow$ mạn        & Med  \\ \hline
139 & 6  & ngày $\Longrightarrow$ này       & -     \\ \hline
140 & 6  & nhổ $\Longrightarrow$ nhỏ        & Med  \\ \hline
141 & 6  & nín $\Longrightarrow$ đến        & Med  \\ \hline
142 & 6  & nó $\Longrightarrow$ là          & Fill \\ \hline
143 & 6  & phải $\Longrightarrow$ cái       & -     \\ \hline
144 & 6  & ra $\Longrightarrow$ da          & -     \\ \hline
145 & 6  & rong $\Longrightarrow$ tâm       & Med  \\ \hline
146 & 6  & sợ $\Longrightarrow$ sở          &  -    \\ \hline
147 & 6  & sữa $\Longrightarrow$ sự         & Med  \\ \hline
148 & 6  & thì $\Longrightarrow$ bị         & Fill \\ \hline
149 & 6  & thì $\Longrightarrow$ chúng      & Fill \\ \hline
150 & 6  & thì $\Longrightarrow$ thể        & Fill \\ \hline
151 & 6  & thú $\Longrightarrow$ thuốc      & Med  \\ \hline
152 & 6  & thấy $\Longrightarrow$ cái       &  -    \\ \hline
153 & 6  & thể $\Longrightarrow$ sẽ         & Med  \\ \hline
154 & 6  & trẻ $\Longrightarrow$ kể         & Med  \\ \hline
155 & 6  & trẻ $\Longrightarrow$ để         & Med  \\ \hline
156 & 6  & trồng $\Longrightarrow$ viêm     & Med  \\ \hline
157 & 6  & u $\Longrightarrow$ ung          & Med  \\ \hline
158 & 6  & viện $\Longrightarrow$ vị        & Med  \\ \hline
159 & 6  & với $\Longrightarrow$ cái        & Fill \\ \hline
160 & 6  & xơ $\Longrightarrow$ thư         & Med  \\ \hline
161 & 6  & âm $\Longrightarrow$ vitamin     & Med  \\ \hline
162 & 6  & đo $\Longrightarrow$ đau         & Med  \\ \hline
163 & 6  & đây $\Longrightarrow$ này        & Fill \\ \hline
164 & 6  & đấy $\Longrightarrow$ đây        & Fill \\ \hline
165 & 6  & đầu $\Longrightarrow$ đau        & Med  \\ \hline
166 & 6  & đầy $\Longrightarrow$ đây        & -     \\ \hline
167 & 6  & đủ $\Longrightarrow$ đúng        &  -    \\ \hline
168 & 5  & cho $\Longrightarrow$ cao        &  -    \\ \hline
169 & 5  & cho $\Longrightarrow$ trong      &  -    \\ \hline
170 & 5  & chân $\Longrightarrow$ nhân      & Med  \\ \hline
171 & 5  & chín $\Longrightarrow$ chính     & Med  \\ \hline
172 & 5  & chỉ $\Longrightarrow$ cái        &  -    \\ \hline
173 & 5  & covid19 $\Longrightarrow$ chính  & Med  \\ \hline
174 & 5  & còn $\Longrightarrow$ và         &  -    \\ \hline
175 & 5  & có $\Longrightarrow$ bác         & Fill \\ \hline
176 & 5  & có $\Longrightarrow$ là          & Fill \\ \hline
177 & 5  & do $\Longrightarrow$ ra          & -     \\ \hline
178 & 5  & dạng $\Longrightarrow$ giảm      &  -    \\ \hline
179 & 5  & dự $\Longrightarrow$ nhiều       &  -    \\ \hline
180 & 5  & gây $\Longrightarrow$ cái        &  -    \\ \hline
181 & 5  & hoặc $\Longrightarrow$ họ        &  -    \\ \hline
182 & 5  & hư $\Longrightarrow$ hơn         & Med  \\ \hline
183 & 5  & không $\Longrightarrow$ trong    & -     \\ \hline
184 & 5  & khỏe $\Longrightarrow$ khoẻ      & Med  \\ \hline
185 & 5  & kinh $\Longrightarrow$ cái       & Med  \\ \hline
186 & 5  & kết $\Longrightarrow$ cái        & Med  \\ \hline
187 & 5  & là $\Longrightarrow$ người       & Fill \\ \hline
188 & 5  & là $\Longrightarrow$ này         & Fill \\ \hline
189 & 5  & là $\Longrightarrow$ đã          & Fill \\ \hline
190 & 5  & mà $\Longrightarrow$ là          & Fill \\ \hline
191 & 5  & mái $\Longrightarrow$ máy        & Med  \\ \hline
192 & 5  & mất $\Longrightarrow$ mức        &  -    \\ \hline
193 & 5  & mặt $\Longrightarrow$ mạch       & Med  \\ \hline
194 & 5  & nang $\Longrightarrow$ năng      &  -    \\ \hline
195 & 5  & nhân $\Longrightarrow$ nhắn      & Med  \\ \hline
196 & 5  & nhũ $\Longrightarrow$ nhiều      & Med  \\ \hline
197 & 5  & này $\Longrightarrow$ ngày       & Fill \\ \hline
198 & 5  & nó $\Longrightarrow$ cái         & Fill \\ \hline
199 & 5  & nó $\Longrightarrow$ có          & Fill \\ \hline
200 & 5  & nền $\Longrightarrow$ nên        & Med  \\ \hline
201 & 5  & phụ $\Longrightarrow$ phẫu       & Med  \\ \hline
202 & 5  & que $\Longrightarrow$ quá        & Med  \\ \hline
203 & 5  & quên $\Longrightarrow$ khuyên    &  -    \\ \hline
204 & 5  & răng $\Longrightarrow$ căn       & Med  \\ \hline
205 & 5  & sao $\Longrightarrow$ ra         &  -    \\ \hline
206 & 5  & sâu $\Longrightarrow$ sau        & Med  \\ \hline
207 & 5  & sẽ $\Longrightarrow$ sĩ          &  -    \\ \hline
208 & 5  & sức $\Longrightarrow$ rất        & Med  \\ \hline
209 & 5  & thanh $\Longrightarrow$ thành    & Med  \\ \hline
210 & 5  & thuyên $\Longrightarrow$ nguyên  & Med  \\ \hline
211 & 5  & thì $\Longrightarrow$ người      & Fill \\ \hline
212 & 5  & thì $\Longrightarrow$ này        & Fill \\ \hline
213 & 5  & thính $\Longrightarrow$ tính     & Med  \\ \hline
214 & 5  & thể $\Longrightarrow$ để         & Med  \\ \hline
215 & 5  & tiêm $\Longrightarrow$ tim       & Med  \\ \hline
216 & 5  & truyền $\Longrightarrow$ trì     & Med  \\ \hline
217 & 5  & tránh $\Longrightarrow$ trình    & Med  \\ \hline
218 & 5  & trên $\Longrightarrow$ chân      & -     \\ \hline
219 & 5  & trắng $\Longrightarrow$ tháng    & Med  \\ \hline
220 & 5  & tức $\Longrightarrow$ rất        &  -    \\ \hline
221 & 5  & tử $\Longrightarrow$ công        & Med  \\ \hline
222 & 5  & và $\Longrightarrow$ giảm        & Fill \\ \hline
223 & 5  & vâng $\Longrightarrow$ vân       & -     \\ \hline
224 & 5  & xơ $\Longrightarrow$ oxy         & Med  \\ \hline
225 & 5  & áp $\Longrightarrow$ tác         & Med  \\ \hline
226 & 5  & âm $\Longrightarrow$ năm         & Med  \\ \hline
227 & 5  & ăn $\Longrightarrow$ anh         & Med  \\ \hline
228 & 5  & đeo $\Longrightarrow$ đều        &  -    \\ \hline
229 & 5  & đâu $\Longrightarrow$ đau        &  -    \\ \hline
230 & 5  & đó $\Longrightarrow$ đã          & -     \\ \hline
231 & 5  & đầu $\Longrightarrow$ nào        & Med  \\ \hline
232 & 5  & để $\Longrightarrow$ thì         &  -    \\ \hline
233 & 5  & để $\Longrightarrow$ đấy         &  -    \\ \hline
234 & 5  & đợt $\Longrightarrow$ được       &  -    \\ \hline
235 & 5  & ở $\Longrightarrow$ của          &  -    \\ \hline
\caption{\label{confusion_pair} Statistics of confusion pairs in \textit{VietMed-Test} using the best pre-trained model \textit{XLSR-53-Viet} (WER = 29.6).
\\ In this table, we divide into 2 types of confusion pairs: Medical (a word that is a part of a medical term) and Filler (a word that is a part of a filler in real-world conversations). 
Only confusion pairs that have at least 5 occurrences in the recognition of the \textit{VietMed-Test} are included in this table.}
\end{longtable}

\newpage

\begin{longtable}{|l|l|c|}
\multicolumn{1}{c}{\textbf{OOV}}           & \multicolumn{1}{c}{\textbf{Phonemes}}                                                 & \textbf{Correct} \\
\endfirsthead
\multicolumn{3}{c}%
{{\bfseries Table \thetable\ continued from previous page}} \\
\multicolumn{1}{c}{\textbf{OOV}}           & \multicolumn{1}{c}{\textbf{Phonemes}}                                                 & \textbf{Correct} \\
\endhead
acenocoumarol & a:\_2 k E\_1 n o\_1 k a\_1 u\_1 m a:\_1 z O\_1 n      & N       \\
alo           & a:\_1 l O\_1                                          & Y       \\
amin          & a:\_1 m i\_1 n                                        & Y       \\
amylase       & a:\_1 m i\_1 l a:\_1                                  & N       \\
apomorphine   & a:\_2 p o\_1 m o\_1 f i\_1 n                          & Y       \\
ascorbic      & a:\_1 s k O\_1 b\_\&amp;lt; i\_2 k                    & Y       \\
aspirin       & a:\_1 s p i\_1 z i\_1 n                               & N       \\
betacarotene  & b\_\&amp;lt; E\_1 t a:\_2 k a:\_1 z O\_1 t E\_1 n     & N       \\
betaglucan    & b\_\&amp;lt; E\_1 t a:\_1 l u\_1 k a:\_1 n            & Y       \\
canxi         & k a:\_1 n s i\_1                                      & Y       \\
catecholamine & k a:\_2 t E\_1 ts\textbackslash O\_1 l a:\_1 m i\_1 n & N       \\
cbt           & k b\_\&amp;lt; t                                      & N       \\
cholesterol   & ts\textbackslash O\_1 l E\_1 s t @:\_1 O\_1 n         & N       \\
clohidric     & k @:\_3 l o\_1 a\_1 z i\_2 k                          & N       \\
collagen      & k o\_1 l l a:\_1 z E\_1 n                             & Y       \\
cologen       & k o\_1 l o\_1 G E\_1 n                                & Y       \\
corticoid     & k O\_1 t i\_1 k O\_1 i\_1                             & Y       \\
cortisol      & k O\_1 t i\_1 s O\_1 n                                & Y       \\
covid         & k o\_1 v i\_1                                         & N       \\
ct            & k t                                                   & N       \\
dbs           & z b\_\&amp;lt;                                        & N       \\
gen           & G E\_1 n                                              & Y       \\
google        & G O\_1 o\_1 G o\_1                                    & N       \\
gút           & G u\_2 t                                              & Y       \\
hdl           & h d\_\&amp;lt; n                                      & N       \\
hemoglobin    & h E\_1 m o\_1 G @:\_3 l O\_1 b\_\&amp;lt; i\_1 n      & Y       \\
hormone       & h O\_1 m O\_1 n                                       & Y       \\
inr           & i\_1 n                                                & N       \\
insulin       & i\_1 n s u\_1 l i\_1 n                                & Y       \\
internet      & i\_1 n t @:\_1 n E\_2 t                               & Y       \\
iod           & i\_1 o\_2 t                                           & Y       \\
kcal          & k k a:\_1 n                                           & N       \\
kilogam       & k i\_1 l o\_1 G a:\_1 m                               & Y       \\
laser         & l a:\_1 @:\_1                                         & N       \\
ldl           & l d\_\&amp;lt; n                                      & N       \\
levodopa      & l E\_1 v o\_1 d\_\&amp;lt; O\_2 p a:\_1               & Y       \\
liraglutide   & l i\_1 z a:\_1 l u\_1 t i\_1 d\_\&amp;lt; E\_1        & N       \\
livestream    & l a:\_1 i\_1 s ts\textbackslash i\_1 m                & Y       \\
mc            & m k                                                   & N       \\
mililit       & m i\_1 l i\_1 l i\_2 t                                & Y       \\
milimet       & m i\_1 l i\_1 m E\_2 t                                & Y       \\
monitor       & m O\_1 n i\_1 t O\_1                                  & Y       \\
mri           & m z i\_1                                              & N       \\
multivitamin  & m u\_1 n t i\_1 v i\_1 t a:\_1 m i\_1 n               & Y       \\
natri         & n a:\_1 t z i\_1                                      & N       \\
niu           & n i\_1 u\_1                                           & Y       \\
noark         & n O\_1 a:\_1 k                                        & Y       \\
orlistat      & O\_1 l i\_2 t a:\_2 t                                 & N       \\
pacemaker     & p a:\_2 k E\_1 m a:\_1 k @:\_1                        & N       \\
parkinson     & p a:\_2 k i\_1 n s O\_1 n                             & N       \\
pepsin        & p E\_2 p s i\_1 n                                     & Y       \\
phytoncide    & f i\_1 t O\_1 n s i\_1 d\_\&amp;lt; E\_1              & N       \\
pp            & p p                                                   & N       \\
protein       & p @:\_3 z o\_1 t i\_1 n                               & N       \\
qr            & k                                                     & N       \\
radiography   & z a:\_1 d\_\&amp;lt; i\_1 o\_1 G @:\_3 z a:\_1 f i\_1 & N       \\
run           & z u\_1 n                                              & N       \\
selen         & s E\_1 l E\_1 n                                       & Y       \\
show          & s @\_1 u\_1                                           & N       \\
sulfonylurea  & s u\_1 l f O\_1 n i\_1 l u\_1 i\_2                    & N       \\
sunfuric      & s u\_1 n f u\_1 i\_2 k                                & N       \\
test          & t E\_2 t                                              & N       \\
umami         & u\_1 m a:\_1 m i\_1                                   & Y       \\
vitamin       & v i\_1 t a:\_1 m i\_1 n                               & Y       \\
vitamina      & v i\_1 t a:\_1 m i\_1 n a:\_1                         & Y       \\
vắcxin        & v a\_2 k s i\_1 n                                     & Y       \\
ôliu          & o\_1 l i\_1 u\_1                                      & Y      
\\
\caption{\label{OOV_IARPA} List of OOVs found in \textit{VietMed-Train}. In this table, only loan words are included together with their corresponding phonemes (in BABEL IARPA format). 
Since the use of the automatic toolkit Sequitur Grapheme-To-Phoneme \cite{G2P_toolkit}, some OOVs are correctly or incorrectly mapped, which we denote as Yes (Y) or No (N).}
\end{longtable}





\end{document}